\documentclass[letterpaper, 10 pt, conference]{ieeeconf}
\IEEEoverridecommandlockouts

\usepackage[colorlinks]{hyperref}
\hypersetup{
    colorlinks,
    linkcolor=black,
    citecolor=black,
    filecolor=magenta,
    urlcolor=cyan,
}

\usepackage{resizegather}

\usepackage{cite}
\usepackage{amsmath,amssymb,amsfonts,mathrsfs}
\usepackage{algorithmic}
\usepackage{graphicx}
\usepackage{textcomp}
\usepackage{xcolor}
\usepackage{tcolorbox}
\usepackage{textcomp}
\usepackage{tablefootnote}
\usepackage{threeparttable}
\usepackage{caption, subcaption}

\usepackage{tikz}
\usepackage{graphicx}
\usepackage{tkz-euclide}

\usetikzlibrary{positioning}
\usetikzlibrary{shapes}
\usetikzlibrary{shapes.misc}
\usetikzlibrary{shapes.geometric}
\usetikzlibrary{plotmarks}
\usetikzlibrary{intersections}
\usetikzlibrary{calc}
\usetikzlibrary{fit}
\usetikzlibrary{patterns,tikzmark}
\usetikzlibrary{matrix,decorations.pathreplacing,calc}

\tikzset{cross/.style={cross out, draw, 
         minimum size=2*(#1-\pgflinewidth), 
         inner sep=0pt, outer sep=0pt}}

\newcommand{\state}[0]{x}
\newcommand{\prel}[0]{p_{\rm{rel}}}
\newcommand{\vrel}[0]{v_{\rm{rel}}}
\newcommand{\preldot}[0]{\dot{p}_{\rm{rel}}}
\newcommand{\vreldot}[0]{\dot{v}_{\rm{rel}}}

\newcommand{\norm}[1]{\left\lVert#1\right\rVert}

\newcommand{\vertiii}[1]{{\left\vert\kern-0.25ex\left\vert\kern-0.25ex\left\vert #1 
    \right\vert\kern-0.25ex\right\vert\kern-0.25ex\right\vert}}

\newtheorem{theorem}{Theorem}

\newtheorem{remark}{Remark}

\newtheorem{definition}{Definition}

\makeatletter
\renewcommand{\fps@figure}{htp}
\renewcommand{\fps@table}{htp}
\makeatother

\def\BibTeX{{\rm B\kern-.05em{\sc i\kern-.025em b}\kern-.08em
    T\kern-.1667em\lower.7ex\hbox{E}\kern-.125emX}}
    
\begin{document}

\title{A Collision Cone Approach for Control Barrier Functions}

\author{Manan Tayal$^{1}$, Bhavya Giri Goswami$^{2}$, Karthik Rajgopal$^{1}$, Rajpal Singh$^{3}$, Tejas Rao$^{1}$, \\Jishnu Keshavan$^{3}$, Pushpak Jagtap$^{1}$, Shishir Kolathaya$^{1}$
\thanks{$^{1}$Robert Bosch Center for Cyber-Physical Systems (RBCCPS), Indian Institute of Science (IISc), Bengaluru.
{\tt\scriptsize \{manantayal, karthikrajgo, tejasmrao, pushpak, shishirk \}@iisc.ac.in} 
.
}%
\thanks{$^{2}$University of Waterloo.
{\tt\scriptsize bhavya.goswami@uwaterloo.ca}
.
}%
\thanks{$^{3}$Department of Mechanical Engineering, Indian Institute of Science (IISc), Bengaluru.
{\tt\scriptsize  \{rajpalsingh, kjishnu \}@iisc.ac.in} 
.
}%
}

\maketitle
\begin{abstract}
This work presents a unified approach for collision avoidance using Collision-Cone Control Barrier Functions (CBFs) in both ground (UGV) and aerial (UAV) unmanned vehicles. We propose a novel CBF formulation inspired by collision cones, to ensure safety by constraining the relative velocity between the vehicle and the obstacle to always point away from each other. The efficacy of this approach is demonstrated through simulations and hardware implementations on the TurtleBot, Stoch-Jeep, and Crazyflie 2.1 quadrotor robot, showcasing its effectiveness in avoiding collisions with dynamic obstacles in both ground and aerial settings. The real-time controller is developed using CBF Quadratic Programs (CBF-QPs). Comparative analysis with the state-of-the-art CBFs highlights the less conservative nature of the proposed approach. Overall, this research contributes to a novel control formation that can give a guarantee for collision avoidance in unmanned vehicles by modifying the control inputs from existing path-planning controllers.
\end{abstract}


\section{Introduction}
\label{section: Introduction}
\par In recent years, advancements in Human-Robot interaction and safety-critical controllers have enabled robots to operate in diverse and challenging environments, with close proximity to humans. Autonomous navigation is one such safety-critical task that has gained considerable attention, due to the rise in deployment of autonomous vehicles and drones in industries. Researchers have developed various control methods for ensuring the safety of such systems which includes artificial potential fields \cite{8022685}\cite{Singletary2021ComparativeAO}, reference governor \cite{6859176}, reachability analysis \cite{8263977} \cite{RA-UAV} \cite{https://doi.org/10.48550/arxiv.2106.13176}, and nonlinear model predictive control \cite{8442967}\cite{yu_mpc_aut_ground_vehicle}.
To establish formal safety guarantees, such as collision avoidance with obstacles, it is essential to employ a safety-critical control algorithm that encompasses both trajectory tracking/planning and prioritizes safety over tracking. One such approach is based on control barrier functions \cite{7040372, Ames_2017} (CBFs), which define a secure state set through inequality constraints and formulate it as a quadratic programming (QP) problem to ensure the forward invariance of these sets over time.

A significant advantage of using CBF-based quadratic programs over other techniques is that they work efficiently on real-time practical applications in complex dynamic environments \cite{Singletary2021ComparativeAO, https://doi.org/10.48550/arxiv.2106.13176}; that is, optimal control inputs can be computed at a very high frequency on off-the-shelf electronics. It can be applied as a fast safety filter over existing path planning controllers\cite{9682803}, making them highly applicable to real-world autonomous systems, including Unmanned Ground Vehicles (UGVs) and Unmanned Aerial Vehicles (UAVs). There are some works which shows the collision avoidance with a static obstacle for point mass models \cite{7040372, https://doi.org/10.48550/arxiv.2103.12382, 9565037}, for UGVs (Unicycle \& Bicycle Model) \cite{XIAO2021109592, https://doi.org/10.48550/arxiv.2106.05341, 9482979, 7864310, 9029446, 9112342} and for UAVs \cite{7525253, 9112342}. 
The Higher Order CBF (HOCBF) based approaches \cite{DBLP:journals/corr/abs-1903-04706, 9516971} is also shown to successfully avoid collisions with static obstacles but lack geometrical intuition. Extension of this framework (HOCBF) for the case of moving obstacles is possible, however, safety guarantees are provided for a subset of the original safe set, thereby making it conservative. We will discuss this in more detail at the end of Section-\ref{section: Safety Guarantee}.

\begin{figure}[t]
    \centering
    \includegraphics[width=0.25\linewidth]{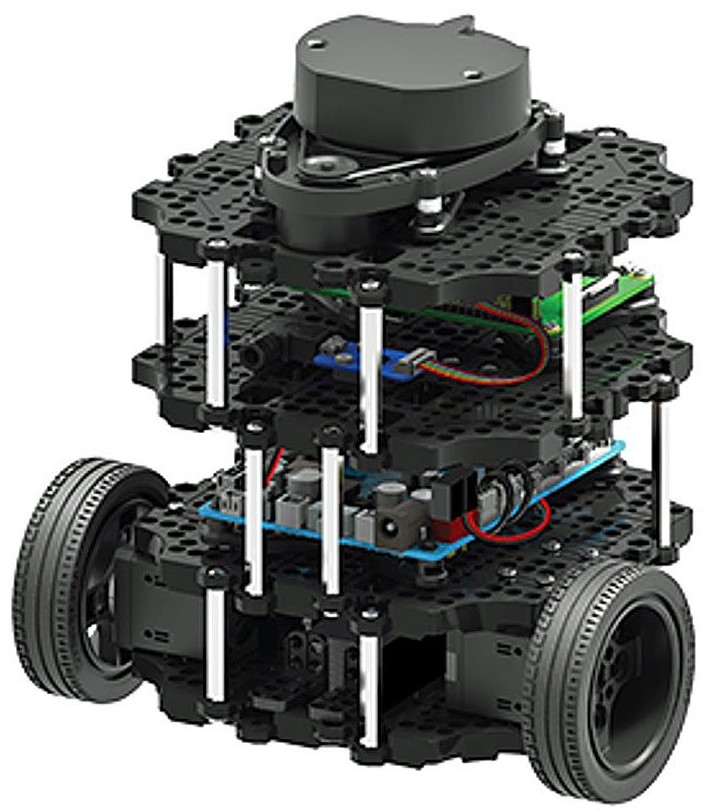}
    \includegraphics[width=0.3\linewidth]{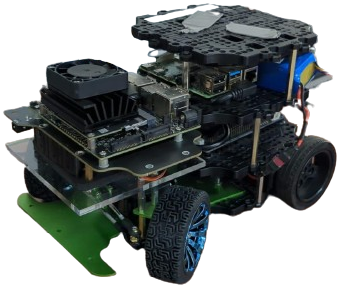}
    \includegraphics[width=0.3\linewidth]{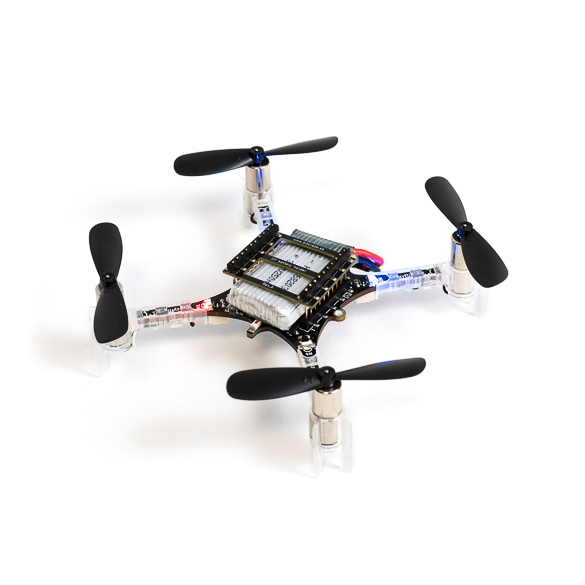}
\caption{Test setups: TurtleBot3 Burger; Stoch-Jeep; Crazyflie 2.1. respectively}
\label{fig:turtlebot stoch-jeep and CF}
\end{figure}

To address the challenges of not handling the dynamic obstacles well and lack of geometric intuition, we propose a new class of CBFs using the concept of collision cones. In particular, we generate a new class of constraints that ensure that the relative velocity between the obstacle and the vehicle always points away from the direction of the vehicle's approach. Thus, giving rise to the \textbf{Collision Cone Control Barrier Functions (C3BFs)} \cite{C3BF_icc, C3BF_UAV, C3BF_UGV_exp}. The C3BF-based QP optimally and rapidly calculates inputs in real-time such that the relative velocity vector direction is kept out of the collision cone for all time. This novel approach enhances the capabilities of collision cones with CBF formulation, allowing them to handle nonholonomic autonomous vehicles and effectively avoid collisions with moving obstacles. The potential of C3BFs is demonstrated using acceleration-controlled unicycle and bicycle models (UGV) and quadrotor model (UAV) (Fig. \ref{fig:turtlebot stoch-jeep and CF}).

The idea of collision cones was first introduced in \cite{Fiorini1993, doi:10.1177/027836499801700706, 709600} as a means to represent the possible set of velocity vectors of the vehicle that lead to a collision. This geometrical approach offers simplicity, efficiency, and adaptability to different environments, making it suitable for a wide range of robotic and autonomous systems. The approach was extended for irregularly shaped robots and obstacles with unknown trajectories both in 2D \cite{709600} and 3D space \cite{Chakravarthy2012}. 
This was commonly used for offline obstacle-free path planning applications \cite{doi:10.2514/1.G005879, doi:10.2514/6.2004-4879} like missile guidance. Collision cones have also been incorporated into Model Predictive Control (MPC) by defining the cones as constraints \cite{babu2018model}, but real-time implementation requires huge computation on board and may face issues like infeasibility of the solution and lack of formal safety guarantees. 

\subsection{Contribution and Paper Structure}
The main idea is to mathematically derive a CBF-QP formulation for the unicycle, bicycle, and quadrotor dynamics for avoiding obstacles with non-zero velocity values.
The main contributions of our work are:

\begin{itemize}
    \item We formulate a direct method for safe trajectory tracking based on collision cone control barrier functions expressed through a quadratic program.
    \item We consider static and constant velocity obstacles of various dimensions and provide mathematical guarantees for collision avoidance.
    \item We compare the collision cone CBF with the state-of-the-art higher order CBF (HOCBF), and show how the former is better in terms of feasibility and safety guarantees. 
\end{itemize}

\subsection{Organisation}
The rest of this paper is organized as follows. Preliminaries explaining the unicycle, bicycle and quadrotor models, the concept of control barrier functions (CBFs) and safety filter designs are introduced in section \ref{section: Background}. Collision Cone CBFs, its application on the above mentioned model to avoid dynamic obstacles and comparision with higher order CBFs (HOCBFs) is discussed in section \ref{section: Safety Guarantee}. The simulation and hardware experimental results will be discussed in section \ref{section: Results}. Finally, we present our conclusion in section \ref{section: Conclusions}. 

\textbf{Notation:} A continuous function $\kappa : [0, d) \rightarrow [0, \infty)$ for some $d > 0$ is said to belong to class-$\mathcal{K}$ if it is strictly increasing and $\alpha(0) = 0$. Here, $d$ is allowed to be $\infty$. The same function can extended to the interval $\kappa: (-b,d)\to (-\infty, \infty)$ with $b>0$ (which is also allowed to be $\infty$), in which case we call it the extended class $\mathcal{K}$ function. $<.\:,\:.>$ denotes the inner product of two vectors.
$\hat{i}$ and $\hat{j}$ stands for unit vectors along x-axis and y-axis respectively. The units used in the simulations are SI units.

\section{Preliminaries}
\label{section: Background}
In this section, first, we will describe the models of unicycle, bicycle and quadrotor. Next, we will formally introduce Control Barrier Functions (CBFs) and their importance for real-time safety-critical control.

\begin{figure}
    \centering
    \includegraphics[width=0.4\linewidth]{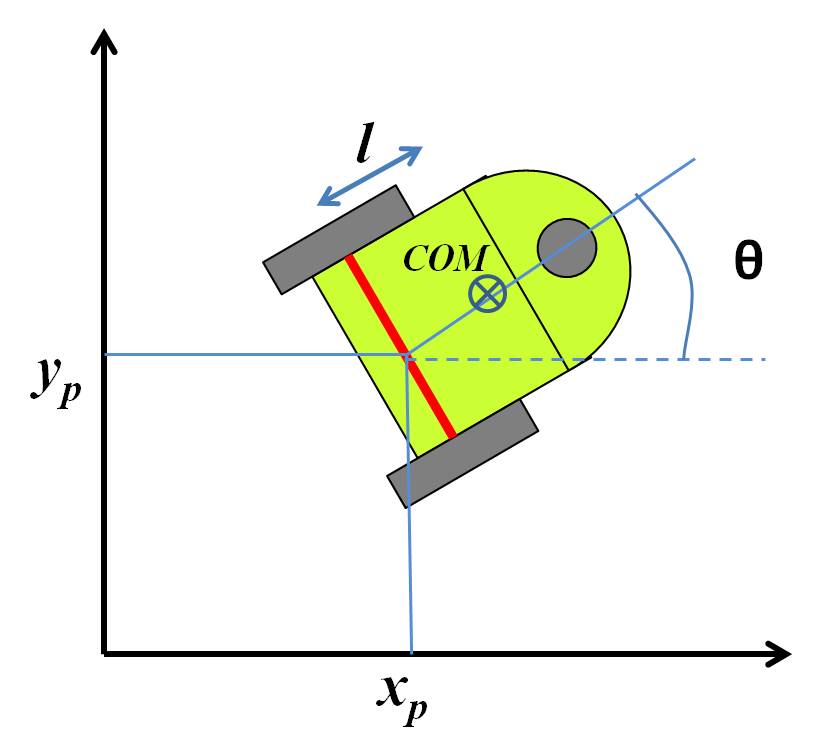}
    \includegraphics[width=0.52\linewidth]{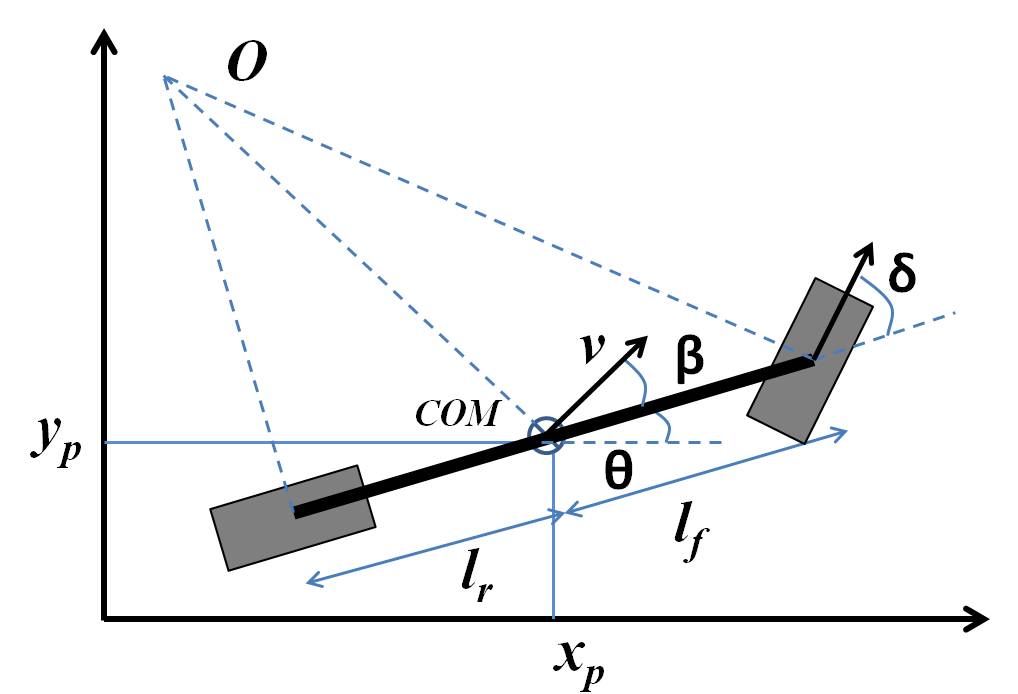}
     \includegraphics[width=0.50\linewidth]{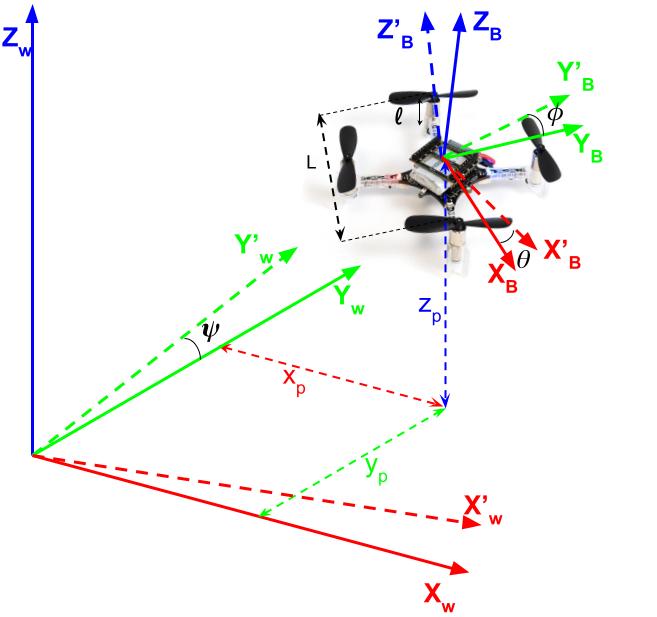}
\caption{Schematic of Unicycle (left); Bicycle model (right); Quadrotor model (down).}
\label{fig:models}
\end{figure}

\subsection{Unicycle model}
\par A unicycle model has state variables $x_p$, $y_p$, $\theta, v, \omega$ denoting the pose, linear velocity, and angular velocity, respectively. The control inputs are linear acceleration $(a)$ and angular  acceleration $(\alpha)$. In Fig. \ref{fig:models} we show a differential drive robot, which is modeled as a unicycle. The resulting dynamics of this model is shown below:

\begin{equation}
	\begin{bmatrix}
		\dot{x}_p \\
		\dot{y}_p \\
		\dot{\theta} \\
		\dot{v} \\
		\dot{\omega}
	\end{bmatrix}
	=
        \begin{bmatrix}
            v\cos\theta\\
            v\sin\theta\\
            \omega \\
            0 \\
            0
        \end{bmatrix}
	+
	\begin{bmatrix}
            0 & 0 \\
            0 & 0 \\
            0 & 0 \\
            1 & 0 \\
            0 & 1
	\end{bmatrix}
	\begin{bmatrix}
		a \\
		\alpha
	\end{bmatrix}
    \label{eqn:Acceleration controlled Unicycle model}
\end{equation}

While the commonly used unicycle model in literature includes linear and angular velocities $v,\omega$ as inputs, we use accelerations as inputs. This is due to the fact that differential drive robots have torques as inputs to the wheels that directly affect accelerations. In other words, we can treat the force / acceleration applied from the wheels as inputs. As a result, $v,\omega$ become state variables in our model.

\subsection{Bicycle model}
The bicycle model has two wheels, where the front wheel is used for steering (see Fig. \ref{fig:models}). This model is typically used for self-driving cars,
where we treat the front and rear wheel sets as a single virtual wheel (for each set) by considering the difference of steer in right and left wheels to be negligible 
\cite{https://doi.org/10.48550/arxiv.2103.12382, Rajamani2014-jb, 7995816}. The bicycle dynamics is as follows:
\begin{align}
    \begin{bmatrix}
        \dot x_p \\
        \dot y_p \\
        \dot \theta \\
        \dot v 
    \end{bmatrix} 
    & = 
    \begin{bmatrix}
        v \cos (\theta + \beta) \\
        v \sin (\theta + \beta) \\
        \frac{v}{l_r} \sin (\beta) \\
        a 
    \end{bmatrix},
    \label{eq:bicyclemodel} \\
	\text{where} \quad \beta &= \tan^{-1}\left(\frac{l_r}{l_f + l_r}\tan(\delta)\right) \label{eq:SlipSteeringConv}
\end{align}
$x_p$ and $y_p$ denote the coordinates of the vehicle’s centre of mass (CoM) in an inertial frame. $\theta$ represents the orientation of the vehicle with respect to the $x$ axis. $a$ is linear acceleration at CoM.
$l_f$ and $l_r$ are the distances of the front and real axle from the CoM,  respectively.
$\delta$ is the steering angle of the vehicle and 
$\beta$ is the vehicle's slip angle, i.e., the steering angle of the vehicle mapped to its CoM (see Fig. \ref{fig:models}).
This is not to be confused with the tire slip angle.

\begin{remark}
Similar to \cite{https://doi.org/10.48550/arxiv.2103.12382}, we assume that the slip angle is constrained to be small. As a result, we approximate $\cos \beta \approx 1$ and $\sin \beta \approx \beta$. Accordingly, we get the following simplified dynamics of the bicycle model:
\begin{equation}
\label{eqn:bicyclemodel with small beta}
	\underbrace{\begin{bmatrix}
		\dot{x}_p \\
		\dot{y}_p \\
		\dot{\theta} \\
		\dot{v}
	\end{bmatrix}}_{\dot{\state}}
	=
	\underbrace{\begin{bmatrix}
		v \cos\theta \\
		v \sin \theta \\
		0 \\
		0
	\end{bmatrix}}_{f(\state)}
	+
	\underbrace{\begin{bmatrix}
		0 & - v\sin\theta \\
		0 & v\cos\theta \\
		0 & \frac{v}{l_r} \\
		1 & 0
	\end{bmatrix}}_{g(\state)}
	\underbrace{\begin{bmatrix}
		a \\
		\beta
	\end{bmatrix}}_{u}.
\end{equation}
Since the control inputs $a, \beta$ are now affine in the dynamics, CBF based Quadratic Programs (CBF-QPs) can be constructed directly to yield real-time control laws, as explained later.
\end{remark}

\subsection{Quadrotor model}
The quadrotor model has four propellers, which provides upward thrusts of $(f_1, f_2, f_3, f_4)$ (see Fig. \ref{fig:models}) and the states needed to describe the quadrotor system is given by $x = [x_p, y_p, z_p, \dot{x}_p, \dot{y}_p, \dot{z}_p,  \phi, \theta, \psi, {\omega}_{1}, {\omega}_{2}, {\omega}_{3}]$ . The quadrotor dynamics is as follows\cite{quadfolk,quan2017introduction}:
\begin{equation}
\label{eqn:quadrotor_model}
	\underbrace{\begin{bmatrix}
		\dot{x}_p \\
		\dot{y}_p \\
            \dot{z}_p \\
            \ddot{x}_p \\
		\ddot{y}_p \\
            \ddot{z}_p \\
		\dot{\phi} \\
            \dot{\theta} \\
            \dot{\psi} \\
		\dot{\omega}_{1} \\
            \dot{\omega}_{2} \\
            \dot{\omega}_{3}
	\end{bmatrix}}_{\dot{\state}}
	=
	\underbrace{\begin{bmatrix}
		\dot{x}_p \\
		\dot{y}_p \\
            \dot{z}_p \\
            0 \\
            0 \\
            -g \\
            \textbf{W}^{-1}
            \begin{bmatrix}
                \omega_{1} \\
                \omega_{2} \\
                \omega_{3}
            \end{bmatrix}\\
            \\
		  -I^{-1} \vec{\omega} \times I \vec{\omega}  \\
            .
	\end{bmatrix}}_{f(\state)}
	+
	\underbrace{\begin{bmatrix}
		\\
            \begin{bmatrix}
                0 & 0 & 0 & 0 \\
                0 & 0 & 0 & 0 \\
                0 & 0 & 0 & 0 
            \end{bmatrix}\\
            \\
            \frac{1}{m}\textbf{R}
            \begin{bmatrix}
                0 & 0 & 0 & 0 \\
                0 & 0 & 0 & 0 \\
                1 & 1 & 1 & 1 
            \end{bmatrix}\\
            \\
            \begin{bmatrix}
                0 & 0 & 0 & 0 \\
                0 & 0 & 0 & 0 \\
                0 & 0 & 0 & 0 
            \end{bmatrix}\\
            \\
		I^{-1}L
            \begin{bmatrix}
                1 & 0 & -1 & 0 \\
                0 & 1 & 0 & -1 \\
                c_{\tau} & -c_{\tau} & c_{\tau} & -c_{\tau} 
            \end{bmatrix}
	\end{bmatrix}}_{g(\state)}
	\underbrace{\begin{bmatrix}
		f_{1} \\
		f_{2} \\
            f_{3} \\
            f_{4} 
	\end{bmatrix}}_{u}
\end{equation}
$x_p$, $y_p$ and $z_p$ denote the coordinates of the base centre of the quadrotor in an inertial frame. $\phi$, $\theta$ and $\psi$ represents the (roll, pitch \& yaw) orientation of the quadrotor.  (see Fig. \ref{fig:models}). \textbf{R} is the rotation matrix (from the body frame to the inertial frame), $m$ is the mass of the quadrotor, $\textbf{W}$ is the transformation matrix for angular velocities from the inertial frame to the body frame, I is the inertia matrix and L is the diagonal length of quadrotor. $c_{\tau}$ is the constant that determines the torque produced by each propeller. 

\subsection{Control barrier functions (CBFs)}
Having described the vehicle models, we now formally introduce Control Barrier Functions (CBFs) and their applications in the context of safety. 
%
Given the unicycle, bicycle and quadrotor model, we have the nonlinear control system in affine form:
\begin{equation}
	\dot{\state} = f(\state) + g(\state)u
	\label{eqn: affine control system}
\end{equation}
where $\state \in \mathcal{D} \subseteq \mathbb{R}^n$ is the state of system, and $u \in \mathbb{U} \subseteq \mathbb{R}^m$ the input for the system. Assume that the functions $f: \mathbb{R}^n \rightarrow \mathbb{R}^n$ and $g: \mathbb{R}^n \rightarrow \mathbb{R}^{n \times m}$ are continuously differentiable. Given a Lipschitz continuous control law $u = k(\state)$, the resulting closed loop system $\dot{\state} = f_{cl}(\state) = f(\state) + g(\state)k(\state)$ yields a solution $\state(t)$, with initial condition $\state(0) = \state_0$.
%
Consider a set $\mathcal{C}$ defined as the \textit{super-level set} of a continuously differentiable function $h:\mathcal{D}\subseteq \mathbb{R}^n \rightarrow \mathbb{R}$ yielding,
\begin{align}
\label{eq:setc1}
	\mathcal{C}                        & = \{ \state \in \mathcal{D} \subset \mathbb{R}^n : h(\state) \geq 0\} \\
\label{eq:setc2}
	\partial\mathcal{C}                & = \{ \state \in \mathcal{D} \subset \mathbb{R}^n : h(\state) = 0\}\\
\label{eq:setc3}
	\text{Int}\left(\mathcal{C}\right) & = \{ \state \in \mathcal{D} \subset \mathbb{R}^n : h(\state) > 0\}
\end{align}
It is assumed that $\text{Int}\left(\mathcal{C}\right)$ is non-empty and $\mathcal{C}$ has no isolated points, i.e. $\text{Int}\left(\mathcal{C}\right) \neq \phi$ and $\overline{\text{Int}\left(\mathcal{C}\right)} = \mathcal{C}$. 
The system is safe w.r.t. the control law $u = k(\state)$ if
	$\forall \: \state(0) \in \mathcal{C} \implies \state(t) \in \mathcal{C} \;\;\; \forall t \geq 0$.
We can mathematically verify if the controller $k(\state)$ is safeguarding or not by using Control Barrier Functions (CBFs), which is defined next.

\begin{definition}[Control barrier function (CBF)]{\it
\label{definition: CBF definition}
Given the set $\mathcal{C}$ defined by \eqref{eq:setc1}-\eqref{eq:setc3}, with $\frac{\partial h}{\partial \state}(\state) \neq 0\; \forall \state \in \partial \mathcal{C}$, the function $h$ is called the control barrier function (CBF) defined on the set $\mathcal{D}$, if there exists an extended \textit{class} $\mathcal{K}$ function $\kappa$ such that for all $\state \in \mathcal{D}$:

\begin{equation}
\begin{aligned}
    \underbrace{\text{sup}}_{ u \in \mathbb{U}}\! \left[\underbrace{\mathcal{L}_{f} h(\state) + \mathcal{L}_g h(\state)u} \iffalse+ \frac{\partial h}{\partial t}\fi_{\dot{h}\left(\state, u\right)} \! + \kappa\left(h(\state)\right)\right] \! \geq \! 0
\end{aligned}
\end{equation}
where $\mathcal{L}_{f} h(\state) = \frac{\partial h}{\partial \state}f(\state)$ and $\mathcal{L}_{g} h(\state)= \frac{\partial h}{\partial \state}g(\state)$ are the Lie derivatives.
}
\end{definition}

Given this definition of a CBF, we know from \cite{Ames_2017} and \cite{8796030} that any Lipschitz continuous control law $k(\state)$ satisfying the inequality: $\dot{h} + \kappa( h )\geq 0$ ensures safety of $\mathcal{C}$ if $x(0)\in \mathcal{C}$, and asymptotic convergence to $\mathcal{C}$ if $x(0)$ is outside of $\mathcal{C}$. 

\subsection{Controller synthesis for real-time safety}
\label{subsection: safe_controller}
Having described the CBF and its associated formal results, we now discuss its Quadratic Programming (QP) formulation. 
CBFs are typically regarded as \textit{safety filters} which take the desired input (reference contoller input) $u_{ref}(\state,t)$ and modify this input in a minimal way: 

\begin{equation}
\begin{aligned}
\label{eqn: CBF-QP}
u^{*}(x,t) &= \min_{u \in \mathbb{U} \subseteq \mathbb{R}^m} \norm{u - u_{ref}(x,t)}^2\\
\quad & \textrm{s.t. } \mathcal{L}_f h(x) + \mathcal{L}_g h(x)u + \kappa \left(h(x)\right) \geq 0\\
\end{aligned}
\end{equation}

This is called the Control Barrier Function based Quadratic Program (CBF-QP). The CBF-QP control $u^{*}$ can be obtained by solving the above optimization problem using KKT conditions.

\subsection{Classical CBFs and moving obstacle avoidance}
Having introduced CBFs, we now explore collision avoidance in Unmanned Ground Vehicles (UGVs). In particular, we discuss the problems associated with the classical CBF-QPs, especially with the velocity obstacles. 
We also summarize and compare with C3BFs in Table \ref{table: types of CBFs}.

\begin{table*}[t]
\caption{Comparison between the Ellipse CBF \eqref{eqn:Ellipse-CBF}, HOCBF \eqref{eq:hocbf} and the proposed C3BF \eqref{eqn:CC-CBF} for different UGV models.}
\begin{center}
    \begin{threeparttable}
	\begin{tabular}{| c | c | c | c |}
		\hline
		\textbf{CBFs} & \textbf{Vehicle Models} & \textbf{Static Obstacle} $(c_x, c_y, c_z)$ & \textbf{Moving Obstacle} $(c_x(t), c_y(t), c_z(t)) ^\dag$ \\
		\hline
		\textbf{Ellipse CBF}  & Unicycle \eqref{eqn:Acceleration controlled Unicycle model}   & Not a valid CBF                        & Not a valid CBF                               \\
		\hline
		\textbf{Ellipse CBF}  & Bicycle \eqref{eqn:bicyclemodel with small beta}    & Valid CBF, No acceleration                        & Not a valid CBF                               \\
  	\hline
		\textbf{Ellipse CBF}  & Quadrotor \eqref{eqn:quadrotor_model}   & Not a valid CBF                        & Not a valid CBF                               \\
		\hline
		\textbf{HOCBF}      & Unicycle  \eqref{eqn:Acceleration controlled Unicycle model}  & Valid CBF, No steering                        & Valid CBF, but conservative  \\
 		\hline
   		\textbf{HOCBF}      & Bicycle \eqref{eqn:bicyclemodel with small beta}  & Valid CBF                        & Not a valid CBF  \\
     \hline
		\textbf{HOCBF}      & Quadrotor \eqref{eqn:quadrotor_model}  & Valid CBF                        & Valid CBF, but conservative  \\
 		\hline
		\textbf{C3BF}   & Unicycle \eqref{eqn:Acceleration controlled Unicycle model} & Valid CBF in $\mathcal{D}$                       & Valid CBF in $\mathcal{D}$                              \\
		\hline
            \textbf{C3BF} & Bicycle \eqref{eqn:bicyclemodel with small beta} & Valid CBF in $\mathcal{C}$ & Valid CBF in $\mathcal{C}$ \\
        \hline
		\textbf{C3BF}   & Quadrotor \eqref{eqn:quadrotor_model} & Valid CBF in $\mathcal{D}$                       & Valid CBF in $\mathcal{D}$                              \\
            \hline
	\end{tabular}
	\begin{tablenotes}\footnotesize
    \item[$\dag$] $(c_x(t), c_y(t), c_z(t))$ are continuous (or at least piece-wise continuous) functions of time
    \end{tablenotes}
	\end{threeparttable}
\end{center}
\label{table: types of CBFs}
\end{table*}

\subsubsection{Ellipse-CBF Candidate}
Consider the following CBF candidate:
\begin{equation}
    h(\state,t) = \left(\frac{c_x(t) - x_p}{c_1}\right)^2 + \left(\frac{c_y(t) - y_p}{c_2}\right)^2 + \left(\frac{c_z(t) - z_p}{c_3}\right)^2 - 1,
    \label{eqn:Ellipse-CBF}
\end{equation}
which approximates an obstacle with an ellipse with center $(c_x(t), c_y(t), c_z(t))$ and axis lengths $c_1,c_2,c_3$. We assume that $c_x(t),c_y(t),c_z(t)$ are differentiable and their derivatives are piece-wise constants. 
Since $h$ in \eqref{eqn:Ellipse-CBF} is dependent on time (e.g., moving obstacles), the resulting set $\mathcal{C}$ is also dependent on time. To analyze this class of sets, time-dependent versions of CBFs can be used \cite{IGARASHI2019735}. Alternatively, we can reformulate our problem to treat the obstacle position $c_x,c_y,c_z$ as states, with their derivatives being constants. This will allow us to continue using the classical CBF given by Definition \ref{definition: CBF definition}, including its properties on safety. The derivative of \eqref{eqn:Ellipse-CBF} is
\begin{align}
\dot h=\frac{2(c_x-x_p)(\dot c_x - v\cos \theta)}{c_1^2} +\frac{2(c_y-y_p)(\dot c_y - v \sin \theta)}{c_2^2},
\end{align}
which has no dependency on the inputs $a, \alpha$. Hence, $h$ will not be a valid CBF for the acceleration-based model \eqref{eqn:Acceleration controlled Unicycle model}. 

However, for static obstacles, if we choose to use the velocity-controlled model (with $v,\omega$ as inputs instead of $a,\alpha$), then $h$ will certainly be a valid CBF, but the vehicle will have limited control capability, i.e., it looses steering $\omega$.

For the bicycle model, the derivative of $h$ \eqref{eqn:Ellipse-CBF} yields
\begin{align}
    \dot h = & 2 (c_x - x_p) ( \dot c_x - v \cos \theta + v (\sin \theta) \beta)/c_1^2 \nonumber \\
    & + 2 (c_y - y_p) ( \dot c_y - v \sin \theta - v (\cos \theta) \beta)/c_2^2,
\end{align}
which only has $\beta$ as the input. Furthermore, the derivatives $\dot c_x, \dot c_y$ are free variables, i.e., the obstacle velocities can be selected in such a way that the constraint $\dot h(x,u)+ \kappa (h(x)) < 0$, whenever $\mathcal{L}_g h=0$. This implies that $h$ is not a valid CBF for moving obstacles for the bicycle model.

For the quadrotor model, the derivative of $h$ \eqref{eqn:Ellipse-CBF} yields
\begin{align}
    \dot h = & 2 (c_x - x_p) ( \dot c_x - \dot{x}_p)/c_1^2  + 2 (c_y - y_p) ( \dot c_y - \dot{y}_p)/c_2^2 \nonumber \\
    & + 2 (c_z - z_p) ( \dot c_z - \dot{z}_p)/c_3^2,
\end{align}
which has no dependency on the inputs $f_1, f_2, f_3, f_4$. Hence, $h$ will not be a valid CBF for the quadrotor model \eqref{eqn:quadrotor_model}. 

\subsubsection{Higher Order CBFs} It is worth mentioning that for the acceleration-controlled nonholonomic models \eqref{eqn:Acceleration controlled Unicycle model}, we can use 
another class of CBFs introduced specifically for constraints with higher relative degrees: HOCBF \cite{DBLP:journals/corr/abs-1903-04706, 9516971} given by:  
\begin{align}
h_2 = \dot h_1 + \kappa(h_1),
\label{eq:hocbf}
\end{align}
where $h_1$ is the equation of ellipse given by \eqref{eqn:Ellipse-CBF}. 
\par Apart from lacking geometrical intuition, $h_2$ for acceleration-controlled unicycle model will result in a conservative, safe set as per \cite[Theorem 3]{9516971}. For acceleration controlled bicycle model \eqref{eq:bicyclemodel} with same HOCBF $h_2$ \eqref{eq:hocbf}, if $\mathcal{L}_{g} h_2 = 0$ then we can choose $\dot c_x , \dot c_y$ in such a way that $\mathcal{L}_f h_2 + \frac{\partial h_2}{\partial t} \ngeq 0$ which results in an invalid CBF. Due to space constraints, a detailed proof of the same is omitted and will be explained as part of future work. The results are summarised in Table-\ref{table: types of CBFs}. 

\par However, our goal in this paper is to develop a CBF formulation with geometrical intuition that provides safety guarantees to avoid moving obstacles with the acceleration-controlled nonholonomic models. We propose this next.


\section{Collision Cone CBF (C3BF)}
\label{section: Safety Guarantee}
A collision cone, defined for a pair of objects, is a set which can be used to predict the possibility of collision between the two objects based on the direction of their relative velocity. Collision cone of an object pair represents the directions, which if traversed by either objects, will result in a collision between the two. 
We will treat the obstacles as ellipses with the vehicle reduced to a point; therefore, throughout the rest of the paper, the term collision cone will refer to this case with the ego-vehicle's center being the point of reference. 





\begin{figure}
    \centering
    \begin{tikzpicture}[
      collisioncone/.style={shape=rectangle, fill=red, line width=2, opacity=0.30},
      obstacleellipse/.style={shape=rectangle, fill=blue, line width=2, opacity=0.35},
    ]
        
        \def\r{1.32003}; 
        \def\q{-3.5}; 
        \def\x{{\r^2/\q}}; 
        \def\y{{\r*sqrt(1-(\r/\q)^2}}; 
        \def\z{{\q - abs(\q - (\r^2/\q))}};
        \coordinate (Q) at (\q,0); 
        \coordinate (P) at (\x,\y); 
        \coordinate (O) at (0.0, 0); 
        \coordinate (E) at (\q, 0); 
        \coordinate (K) at (\x, {-\y}); 
        \coordinate (H) at (\z, \y);
        \coordinate (I) at (\z, {-\y});
        
        \draw[name path = aux, red!60, very thick, dashed] (O) circle (1.32003);
        \draw[blue!50, thick, fill=blue!20] (O) ellipse (1.20 and 0.55);
        \draw[black, thick] (E) -- (O) node [midway, below] {$\|\prel\|$};
        
        
        \draw[black, thick, name path = tangent] ($(Q)!-0.0!(P)$) -- ($(Q)!1.3!(P)$); 
        \draw[black, thick, name path = normal] ($(O)!-0.0!(P)$) -- ($(O)!1.4!(P)$);
        \draw[black, thick] ($(Q)!-0.0!(K)$) -- ($(Q)!1.3!(K)$);
        
        \draw[black, thick, name path = tangent, dashed] ($(Q)!-0.0!(H)$) -- ($(Q)!1.1!(H)$);
        \draw[black, thick, dashed] ($(Q)!-0.0!(I)$) -- ($(Q)!1.1!(I)$);
        
        
        \tkzMarkRightAngle[draw=black,size=.2](O,P,Q);
        \tkzMarkAngle[draw=black, size=0.75](O,Q,P);
        \tkzLabelAngle[dist=1.0](O,Q,P){$\phi$};
        
        \path[shade, left color=red, right color = red, opacity=0.2] (E) -- (H) -- (I) -- cycle;
        
        \fill [black] (E) circle (2pt) node[anchor=north, black] (n1) {$(x,y)$} node[anchor=south, black] (n1) {E};
        \fill [blue] (O) circle (2pt) node[anchor=north, blue] (n1) {$(c_x, c_y)$} node[anchor=south east, blue] (n1) {O}; 
        \fill [black] (P) circle (2pt) node[anchor=south, black] (n1) {$P$};
        \fill [black] (K) circle (2pt) node[anchor=north, black] (n1) {$K$};
        \fill [black] (H) circle (2pt) node[anchor=south, black] (n2) {$H$};
        \fill [black] (I) circle (2pt) node[anchor=north, black] (n1) {$I$};
        
        \draw [<->, color=black, thick, dashed] ([xshift=5 pt, yshift=0 pt]O) -- ([xshift=5 pt, yshift=0 pt]P) node [midway, right] {$r = a+\frac{w}{2}$};
        \draw [<->, color=black, thick, dashed] (O) -- (1.20, 0) node [midway, above] {$a$};
        
        \matrix [above right,nodes in empty cells, matrix of nodes, column sep=0.5cm, inner sep=6pt] at (current bounding box.north west) {
          \node [collisioncone,label=right:{\footnotesize Collision Cone}] {}; &
          \node [obstacleellipse,label=right:{\footnotesize Obstacle Ellipse}] {}; \\
        };
    \end{tikzpicture}
    \caption{Construction of collision cone for an elliptical obstacle considering the ego-vehicle's dimensions (width: $w$).} 
    \label{Fig:Construction of Collision Cone}
\end{figure}
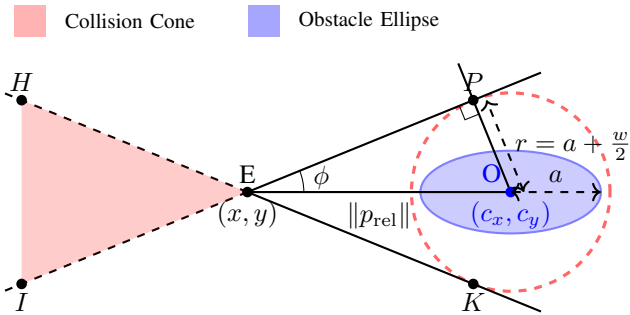

Consider an ego-vehicle defined by the system (\ref{eqn: affine control system}) and a moving obstacle (pedestrian, another vehicle, etc.). This is shown pictorially in Fig. \ref{Fig:Construction of Collision Cone}. We define the velocity and positions of the obstacle w.r.t. the ego vehicle. We \textit{over-approximated} the obstacle to be an ellipse and draw two tangents from the vehicle's centre to a conservative circle encompassing the ellipse, taking into account the ego-vehicle's dimensions ($r = max(c_1, c_2) + \frac{Width_{vehicle}}{2}$). 
For a collision to happen, the relative velocity of the obstacle must be pointing towards the vehicle. Hence, the relative velocity vector must not be pointing into the pink shaded region EHI in Fig. \ref{Fig:Construction of Collision Cone}, which is a cone. 
%
%
Let $\mathcal{C}$ be this set of safe directions for this relative velocity vector. If there exists a function $h:\mathcal{D}\subseteq \mathbb{R}^n \rightarrow \mathbb{R}$ satisfying \textit{Definition: \ref{definition: CBF definition}} on $\mathcal{C}$, then we know that a Lipshitz continuous control law obtained from the resulting QP (\ref{eqn: CBF-QP}) for the system ensures that the vehicle won't collide with the obstacle even if the reference $u_{ref}$ tries to direct them towards a collision course. 
%
%
%
This novel approach of avoiding the pink cone region
gives rise to \textbf{Collision Cone Control Barrier Functions (C3BFs)}. 
We propose the following CBF candidate:
%
\begin{equation}
    h(\state) = <\prel,\vrel> + \|\prel\|\|\vrel\|\cos\phi
    \label{eqn:CC-CBF}
\end{equation}
where, $\prel$ is the relative position vector between the body center of the vehicle and the center of the obstacle, $\vrel$ is the relative velocity, $\phi$ is the half angle of the cone, the expression of $\cos\phi$ is given by $\frac{\sqrt{\|\prel\|^2 - r^2}}{\|\prel\|}$ (see Fig. \ref{Fig:Construction of Collision Cone}). 
The constraint simply ensures that the angle between $\prel, \vrel$ is less than $180^\circ - \phi$.  

\subsection{Application to Unicycle}
\label{section: unicycle model cccbf}
We first obtain the relative position vector between the body center of the unicycle and the center of the obstacle. 
Therefore, we have
\begin{align}\label{eq:positionvectorunicycle}
    \prel := \begin{bmatrix}
        c_x - (x_p + l \cos(\theta)) \\
        c_y - (y_p + l \sin(\theta))
    \end{bmatrix}
\end{align}
Here $l$ is the distance of the body center from the differential drive axis (see Fig. \ref{fig:models}). We obtain its velocity as
\begin{align}\label{eq:velocityvectorunicycle}
    \vrel := \begin{bmatrix}
        \dot c_x - (v \cos (\theta) - l \sin(\theta)*\omega) \\
        \dot c_y - (v \sin (\theta) + l \cos(\theta)*\omega)
    \end{bmatrix}.
\end{align}

We have the following first result of the paper:
%
\begin{theorem}\label{thm:unicycletheorem}{\it
Given the acceleration controlled unicycle model \eqref{eqn:Acceleration controlled Unicycle model}, the proposed CBF candidate \eqref{eqn:CC-CBF} with $\prel,\vrel$ defined by \eqref{eq:positionvectorunicycle}, \eqref{eq:velocityvectorunicycle} is a valid CBF defined for the set $\mathcal{D}$.}
\end{theorem}
\begin{proof}
Taking the derivative of \eqref{eqn:CC-CBF} yields
\begin{align}
\dot h = &  < \preldot, \vrel > + < \prel, \vreldot >  \nonumber \\
 & + < \vrel, \vreldot > \frac{\sqrt{\|\prel\|^2 - r^2}}{\|\vrel\|} \nonumber \\
 & + < \prel, \preldot > \frac{\|\vrel\| }{\sqrt{\|\prel\|^2 - r^2}}.
 \label{eqn:h_derivative}
\end{align}
Further $\preldot  = \vrel$ and
\begin{align*}
    \vreldot = \begin{bmatrix}
        - a \cos \theta + v (\sin \theta) \omega + l (\cos \theta) \omega^2 + l (\sin \theta) \alpha \\
        -a \sin \theta - v (\cos \theta) \omega + l (\sin \theta) \omega^2 - l (\cos \theta) \alpha
    \end{bmatrix}. \nonumber
\end{align*}
Given $\vreldot$ and $\dot h$, we have the following expression for $\mathcal{L}_g h$:
\begin{align}
    \mathcal{L}_g h = \begin{bmatrix}
        < \prel + \vrel \frac{\sqrt{\|\prel\|^2 - r^2}}{\|\vrel\|}, \begin{bmatrix}
            - \cos \theta \\
            - \sin \theta
        \end{bmatrix}>  \\
                < \prel + \vrel \frac{\sqrt{\|\vrel\|^2 - r^2}}{\|\vrel\|}, \begin{bmatrix}
            l \sin \theta \\
            - l \cos \theta
        \end{bmatrix}> 
    \end{bmatrix}^T,
\end{align}
It can be verified that for $\mathcal{L}_gh$ to be zero, we can have the following scenarios:
\begin{itemize}
    \item $\prel + \vrel \frac{\sqrt{\|\prel\|^2 - r^2}}{\|\vrel\|}=0$, which is not possible. Firstly, $\prel=0$ indicates that the vehicle is already inside the obstacle. Secondly, if the above equation were to be true for a non-zero $\prel$, then $\vrel/\|\vrel\| = - \prel/\sqrt{\|\prel\|^2 - r^2}$. This is also not possible as the magnitude of LHS is $1$, while that of RHS is $>1$.
    \item $\prel + \vrel \frac{\sqrt{\|\vrel\|^2 - r^2}}{\|\vrel\|}$ is perpendicular to both $\begin{bmatrix}
            - \cos \theta \\
            - \sin \theta
        \end{bmatrix}$ and  $\begin{bmatrix}
            l \sin \theta \\
            - l \cos \theta
        \end{bmatrix}$, which is also not possible.
\end{itemize}
This implies that $\mathcal{L}_gh$ is always a non-zero matrix, implying that $h$ is a valid CBF.
\end{proof}
\begin{remark}
{\it
Since $\mathcal{L}_g h \neq 0$, we can infer from \cite[Theorem 8]{XU201554} that the resulting QP given by \eqref{eqn: CBF-QP} is Lipschitz continuous. Hence, we can construct CBF-QPs with the proposed CBF \eqref{eqn:CC-CBF} for the unicycle model and guarantee collision avoidance. In addition,  if $h(x(0))<0$, then we can construct a class $\mathcal{K}$ function $\kappa$ in such a way that the magnitude of $h$ exponentially decreases over time, thereby minimizing the violation. We will demonstrate these scenarios in 
Section \ref{section: Results}.
}
\end{remark}

\subsection{Application to Bicycle}
\label{section: bicycle model cccbf}

For the approximated bicycle model \eqref{eqn:bicyclemodel with small beta}, we define the following:
\begin{align}\label{eq:positionvectorbicycle}
    \prel := &
    \begin{bmatrix}
        c_x - x_p &       c_y - y_p 
    \end{bmatrix}^T \\
\label{eq:velocityvectorbicycle}
 \vrel := & \begin{bmatrix}
        \dot c_x - v \cos \theta &        \dot c_y - v \sin \theta 
    \end{bmatrix}^T,
\end{align}

Here $\vrel$ is NOT equal to the relative velocity $\preldot$. However, for small $\beta$, we can assume that $\vrel$ is the difference between obstacle velocity and the velocity component along the length of the vehicle $v \cos \beta \approx v$. In other words, the goal is to ensure that this approximated velocity $\vrel$ is pointing away from the cone. This is an acceptable approximation as $\beta$ is small and the obstacle radius chosen was conservative (see Fig. \ref{Fig:Construction of Collision Cone}). 

We have the following result:
\begin{theorem}\label{thm:bicycletheorem}{\it
Given the bicycle model \eqref{eqn:bicyclemodel with small beta}, the proposed candidate CBF \eqref{eqn:CC-CBF} with $\prel,\vrel$ defined by \eqref{eq:positionvectorbicycle}, \eqref{eq:velocityvectorbicycle} is a valid CBF defined for the set $\mathcal{C}$.}
\end{theorem}
\begin{proof}
We need to show that $\mathcal{L}_g h = 0$ $\implies$ $\dot h + \kappa(h)\geq0$. The derivative of $h$ \eqref{eqn:CC-CBF} yields
\eqref{eqn:h_derivative}. Further using \eqref{eqn:bicyclemodel with small beta}: $\preldot = \vrel + \beta[ v \sin \theta, - v \cos \theta]^T$ and 
\begin{align}
    \vreldot = \begin{bmatrix} 
        -  \cos \theta &  v\sin \theta  \\
        -  \sin \theta & - v \cos \theta 
    \end{bmatrix}\begin{bmatrix}
        a \\ \frac{v}{l_r}\beta
    \end{bmatrix}.
\end{align}

When $\mathcal{L}_g h=0$, we have 

\begin{equation}
    \dot h + \kappa (h) = < \vrel, \vrel > +  \frac{< \prel, \vrel >\|\vrel\| }{\sqrt{\|\prel\|^2 - r^2}} + \kappa (h). \nonumber
\end{equation}
Rewriting the above equation yields
\begin{equation}
    \frac{\|\vrel\|}{\sqrt{\|\prel\|^2 - r^2}}\left( h + \frac{ \sqrt{\|\prel\|^2 - r^2}}{\|\vrel\|}\kappa (h)
    \right).
\end{equation}  
Since $\sqrt{\|\prel\|^2 - r^2}$ and $\|\vrel\|$ are positive quantities, the entire quantity above is $\geq 0$ for all $x\in \mathcal{C}$. This completes the proof. 
\end{proof}
\begin{remark}
{\it
Theorem \ref{thm:bicycletheorem} is different from Theorem \ref{thm:unicycletheorem} as the CBF inuquality is satisfied in the set $\mathcal{C}$ and not in $\mathcal{D}$. In other words, forward invariance of $\mathcal{C}$ can be guaranteed, but not asymptotic convergence of $\mathcal{C}$. However modifications of the control formulation is possible to extend the result for $\mathcal{D}$, which will a subject of future work.
}
\end{remark}

\subsection{Application to Quadrotor}
\begin{figure}[t]
    \includegraphics[width=0.9\linewidth]{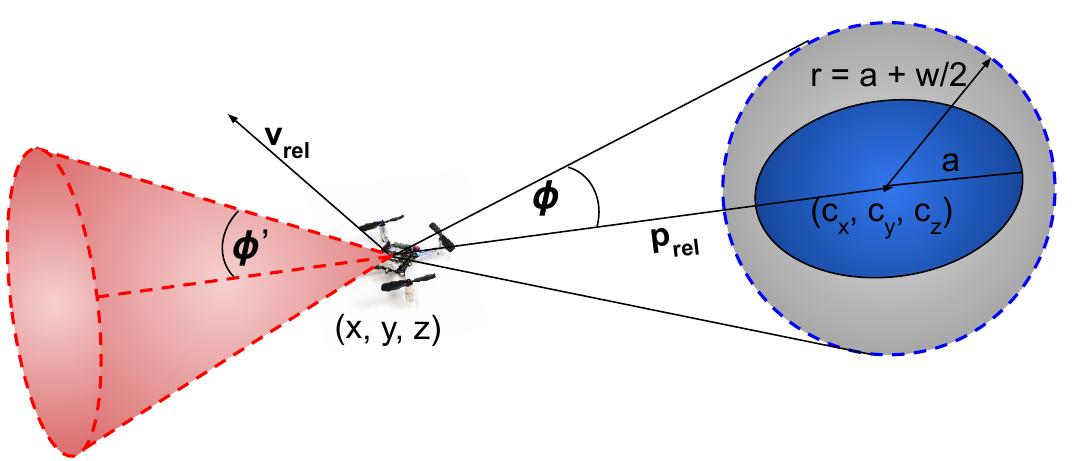}
\caption{\textbf{3D CBF} candidate: The dimensions of the obstacle are comparable to each other, it can be assumed as a sphere}
\label{fig:3D CBF}
\end{figure}

Based on the shape of the obstacle we can divide the proposed candidates into two cases: 1) When the dimensions of the obstacle are comparable to each other, we can assume the obstacle as a sphere with radius $r = max(c_1, c_2, c_3) + \frac{w}{2}$, where $w$ is the max width of the quadrotor absorbed in the obstacle width (shown in Fig. \ref{fig:3D CBF}). We call the CBF candidate so formed in this case as \textbf{3D CBF} candidate (see Fig. \ref{fig:3D CBF}). 2) When one of the dimensions is far bigger than the other dimensions, we can assume the obstacle as a cylinder with height $H = max(c_1, c_2, c_3) $ and radius $r = max_{2}(c_1, c_2, c_3) + \frac{w}{2}$ (where $max_2$ is the second largest element in the list) and we call the candidate so formed in this case as \textbf{Projection CBF} candidate (see Fig. \ref{fig:Projection CBF}).


\subsubsection{3D CBF candidate}
\label{section: 3D-CBF}
When the dimensions of the obstacle are comparable to each other, we can assume the obstacle as a sphere with radius $r = max(c_1, c_2, c_3) + \frac{w}{2}$, where $w$ is the max width of the quadrotor absorbed in the obstacle width (shown in Fig. \ref{fig:3D CBF}). We call the CBF candidate so formed in this case as \textbf{3D CBF} candidate (see Fig. \ref{fig:3D CBF}). We first obtain the relative position vector between the body center of the quadrotor and the center of the obstacle. 
We have the following:
\begin{align}\label{eq:pos-vec-3D}
    \prel := \begin{bmatrix}
        c_x \\
        c_y \\
        c_z
    \end{bmatrix}
    - \left (
    \begin{bmatrix}
        x_p \\
        y_p \\
        z_p
    \end{bmatrix}
    + \textbf{R} \begin{bmatrix}
        0 \\
        0 \\
        l
    \end{bmatrix}
    \right )
\end{align}
Here $l$ is the distance of the body center from the base (see Fig. \ref{fig:models}). $c_x,c_y,c_z$ represents the obstacle location as a function of time. Also, since the obstacles are of constant velocity, we have $\Ddot{c}_x= \Ddot{c}_y= \Ddot{c}_z = 0$. We obtain its relative velocity as
\begin{align}\label{eq:vel-vec-3D}
    \vrel := \dot{p}_{rel}
\end{align}

Now, we calculate the $\vreldot$ term which contains our inputs i.e. $(f_1, f_2, f_3, f_4)$, as follows:

\begin{equation}\label{eqn:vrel-dot-3D}
    \vreldot = -\textbf{R}
            \begin{bmatrix}
                0 & \frac{Ll}{I_{yy}} & 0 & \frac{-Ll}{I_{yy}} \\
                \frac{-Ll}{I_{xx}} & 0 & \frac{Ll}{I_{xx}} & 0 \\
                \frac{1}{m} & \frac{1}{m} & \frac{1}{m} & \frac{1}{m} 
            \end{bmatrix}
            \begin{bmatrix}
    		f_{1} \\
    		f_{2} \\
                f_{3} \\
                f_{4} 
    	\end{bmatrix} \\
        + \rm{additional} \: \rm{terms}. \nonumber
\end{equation}

We have the following result.

\begin{theorem}\label{thm:CC-CBF-3D}{\it
Given the quadrotor model \eqref{eqn:quadrotor_model}, the proposed CBF candidate \eqref{eqn:CC-CBF} with $\prel,\vrel$ defined by \eqref{eq:pos-vec-3D}, \eqref{eq:vel-vec-3D} is a valid CBF defined for the set $\mathcal{D}$.}
\end{theorem}
\begin{proof}
Taking the derivative of \eqref{eqn:CC-CBF} yields
\begin{align}
\dot h = &  < \preldot, \vrel > + < \prel, \vreldot >  \nonumber \\
 & + < \vrel, \vreldot > \frac{\sqrt{\|\prel\|^2 - r^2}}{\|\vrel\|} \nonumber \\
 & + < \prel, \preldot > \frac{\|\vrel\| }{\sqrt{\|\prel\|^2 - r^2}}.
 \label{eqn:CC_h_d}
\end{align}
By substituting for $\vreldot$ (which contains the input) in $\dot h$ \eqref{eqn:CC_h_d}, we have the following expression for $\mathcal{L}_g h$:
\begin{align}
    \mathcal{L}_g h = \begin{bmatrix}
        < \prel + \vrel \frac{\sqrt{\|\prel\|^2 - r^2}}{\|\vrel\|}, 
            \textbf{R}\begin{bmatrix}
                0  \\
                \frac{-Ll}{I_{xx}}\\
                \frac{1}{m}
            \end{bmatrix}>\\
                < \prel + \vrel \frac{\sqrt{\|\vrel\|^2 - r^2}}{\|\vrel\|}, 
            \textbf{R}\begin{bmatrix}
                \frac{Ll}{I_{yy}} \\
                0\\
                \frac{1}{m}
            \end{bmatrix}>\\
         < \prel + \vrel \frac{\sqrt{\|\vrel\|^2 - r^2}}{\|\vrel\|}, 
            \textbf{R}\begin{bmatrix}
                0  \\
                \frac{Ll}{I_{xx}}\\
                \frac{1}{m}
            \end{bmatrix}>\\
         < \prel + \vrel \frac{\sqrt{\|\vrel\|^2 - r^2}}{\|\vrel\|}, 
            \textbf{R}\begin{bmatrix}
                \frac{-Ll}{I_{yy}} \\
                0 \\
                \frac{1}{m}
            \end{bmatrix}>
    \end{bmatrix}^T,
\end{align}

It can be verified that for $\mathcal{L}_gh$ to be zero, we can have the following scenarios:
\begin{itemize}
    \item $\prel + \vrel \frac{\sqrt{\|\prel\|^2 - r^2}}{\|\vrel\|}=0$, which is not possible. Firstly, $\prel=0$ indicates that the vehicle is already inside the obstacle. Secondly, if the above equation were to be true for a non-zero $\prel$, then $\vrel/\|\vrel\| = - \prel/\sqrt{\|\prel\|^2 - r^2}$. This is also not possible as the magnitude of LHS is $1$, while that of RHS is $>1$.
    \item $\prel + \vrel \frac{\sqrt{\|\vrel\|^2 - r^2}}{\|\vrel\|}$ is perpendicular to all  $\textbf{R}\begin{bmatrix}
                    0  \\
                    \frac{-Ll}{I_{xx}}\\
                    \frac{1}{m}
                \end{bmatrix}$, 
    $\textbf{R}\begin{bmatrix}
                \frac{Ll}{I_{yy}} \\
                0\\
                \frac{1}{m}
            \end{bmatrix} $,
    $\textbf{R}\begin{bmatrix}
                0  \\
                \frac{Ll}{I_{xx}}\\
                \frac{1}{m}
            \end{bmatrix} $
            and  $\textbf{R}\begin{bmatrix}
                \frac{-Ll}{I_{yy}} \\
                0 \\
                \frac{1}{m}
            \end{bmatrix} $, which is also not possible. (Because three of these vectors form basis vectors for $\mathbb{R}^{3}$)
\end{itemize}
This implies that $\mathcal{L}_gh$ is always a non-zero matrix, implying that $h$ is a valid CBF.
\end{proof}

\subsubsection{Projection CBF candidate}
\label{section: proj-CBF}

\begin{figure}[t]
    \centering
    \includegraphics[width=0.90\linewidth]{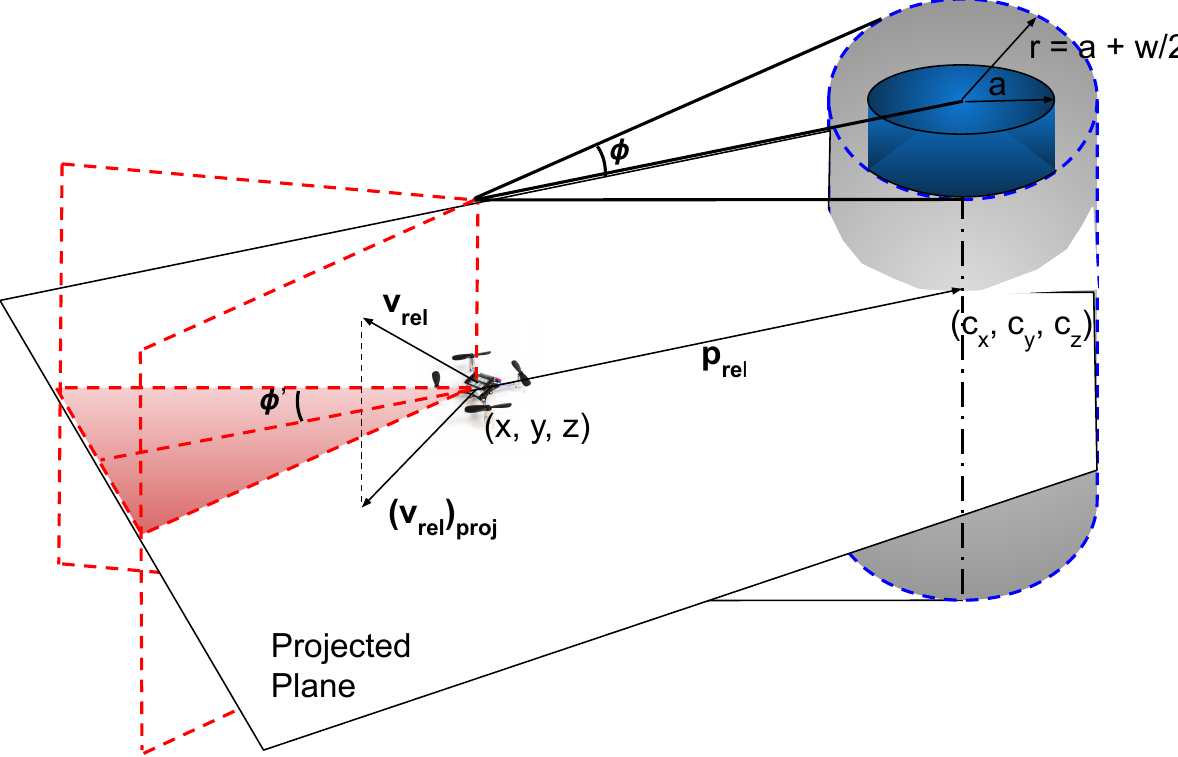}
\caption{\textbf{Projection CBF} candidate: One of the dimensions, of the obstacle, is bigger than the other dimensions, it can be assumed as a cylinder.}
\label{fig:Projection CBF}
\end{figure}

When one of the dimensions is far bigger than the other dimensions, we can assume the obstacle as a cylinder with height $H = max(c_1, c_2, c_3) $ and radius $r = max_{2}(c_1, c_2, c_3) + \frac{w}{2}$ (where $max_2$ is the second largest element in the list) and we call the candidate so formed in this case as \textbf{Projection CBF} candidate (see Fig. \ref{fig:Projection CBF}). 
Now, we have to consider the collision avoidance for elongated (cylindrical) obstacles. We have to obtain the relative position vector between the body center of the quadrotor and the intersection of the axis of the obstacle and the projection plane, where the projection plane is the plane perpendicular to the axis. Therefore, we have

\begin{align}\label{eqn:pos-vec-proj}
    (\prel)_{proj} := \mathcal{P}\left (\begin{bmatrix}
        c_x \\
        c_y \\
        c_z
    \end{bmatrix}
    - \left (
    \begin{bmatrix}
        x_p \\
        y_p \\
        z_p
    \end{bmatrix}
    + \textbf{R} \begin{bmatrix}
        0 \\
        0 \\
        l
    \end{bmatrix}
    \right ) \right ).
\end{align}
Here $l$ is the distance of the body center from the base (see Fig. \ref{fig:models}). $\mathcal{P}: \mathbb{R}^3 \to \mathbb{R}^3 $ is the projection operator, which can be assumed to be a constant\footnote{Note that the obstacles are always translating and not rotating. In addition, it is not restrictive to assume that the translation direction is always perpendicular to the cylinder axis. This makes the projection operator a constant.}. 
Now, since the relative position lies on the projection plane, we have one more condition to satisfy:
\begin{align}\label{eqn:p_in_plane}
    < (\prel)_{proj}, \hat{n} > = 0,   
\end{align}
where, $\hat{n}$ is the normal to the plane. Also, the relative velocity is given by:
\begin{align}\label{eqn:vel-vec-proj}
    (\vrel)_{proj} := \frac{d({p}_{rel})_{proj}}{dt} = (\preldot)_{proj}
\end{align}
Now, we calculate the $\frac{d}{dt}(\vrel)_{proj}$ term which contains our inputs i.e. $(f_1, f_2, f_3, f_4)$, as follows:

\begin{equation}
    \frac{d}{dt}(\vrel)_{proj} =
    \mathcal{P}(
            -  
            \textbf{R}
            \begin{bmatrix}
                0 & \frac{Ll}{I_{yy}} & 0 & \frac{-Ll}{I_{yy}} \\
                \frac{-Ll}{I_{xx}} & 0 & \frac{Ll}{I_{xx}} & 0 \\
                \frac{1}{m} & \frac{1}{m} & \frac{1}{m} & \frac{1}{m} 
            \end{bmatrix}
            \begin{bmatrix}
    		f_{1} \\
    		f_{2} \\
                f_{3} \\
                f_{4} 
    	\end{bmatrix} \\
        + \rm{additional} \: \rm{terms}). \nonumber
\end{equation}
or, from \eqref{eqn:vrel-dot-3D}, we have
\begin{equation}\label{eqn:vrel_dot}
    \frac{d}{dt}(\vrel)_{proj} =
    \mathcal{P}( 
            \vreldot )
\end{equation}
$\frac{d}{dt}(\vrel)_{proj}$  is the projection of $\vreldot$ in \eqref{eqn:vrel-dot-3D} on the projection plane, that is:
\begin{equation}
\begin{aligned}\label{eqn:vrel-dot-proj}
    (\vreldot)_{proj} = \vreldot - <\vreldot, \hat{n}> \hat{n}.
\end{aligned}
\end{equation}

Thus, from \eqref{eqn:p_in_plane} and \eqref{eqn:vrel-dot-proj}, we have the following:
\begin{equation}
\begin{aligned}\label{eqn:pdot-vreldot}
    <(\prel)_{proj}, (\vreldot)_{proj}> = <(\prel)_{proj}, \vreldot>.
\end{aligned}
\end{equation}

Similarly,
\begin{equation}
\begin{aligned}\label{eqn:vdot-vreldot}
    <(\vrel)_{proj}, (\vreldot)_{proj}> = <(\vrel)_{proj}, \vreldot>
\end{aligned}
\end{equation}

We now provide the formal results for the Projection case of \eqref{eqn:CC-CBF} in this subsection. The candidate is given as follows:
\begin{equation}
\begin{aligned}
    h_{proj}(\state, t) = < (\prel)_{proj}, (\vrel)_{proj}> + \\
    \| (\prel)_{proj}\|\| (\vrel)_{proj}\|\cos\phi
    \label{eqn:CC-CBF-proj}
\end{aligned}
\end{equation}
where, $\phi$ is the half angle of the cone, the expression of $\cos\phi$ is given by $\frac{\sqrt{\|(\prel)_{proj}\|^2 - r^2}}{\|(\prel)_{proj}\|}$ (see Fig. \ref{fig:Projection CBF}).  We now show that the proposed CBF candidate \eqref{eqn:CC-CBF-proj} is indeed a valid CBF. 

\begin{theorem}\label{thm:CC-CBF-proj}{\it
Given the quadrotor model \eqref{eqn:quadrotor_model}, the proposed CBF candidate \eqref{eqn:CC-CBF-proj} with $\prel,\vrel$ defined by \eqref{eqn:pos-vec-proj}, \eqref{eqn:vel-vec-proj} is a valid CBF defined for the set $\mathcal{D}$.}
\end{theorem}
\begin{proof}
We have the following derivative of $h_{proj}$:
\begin{align}
\dot h_{proj} = &  < (\preldot)_{proj}, (\vrel)_{proj} > + < (\prel)_{proj}, (\vreldot)_{proj} >  \nonumber \\
 & + < (\vrel)_{proj}, (\vreldot)_{proj} > \frac{\sqrt{\|(\prel)_{proj}\|^2 - r^2}}{\|(\vrel)_{proj}\|} \nonumber \\
 & + < (\prel)_{proj}, (\preldot)_{proj} > \frac{\|(\vrel)_{proj}\| }{\sqrt{\|(\prel)_{proj}\|^2 - r^2}}.
 \label{eqn:CC_h_d-proj}
\end{align}
 $\vreldot$ (which contains the input) from \eqref{eqn:vrel-dot-proj}, equations \eqref{eqn:pdot-vreldot}, \eqref{eqn:vdot-vreldot} and $\dot h_{proj}$ \eqref{eqn:CC_h_d-proj}, we have the following expression for $\mathcal{L}_g h_{proj}$:
\begin{align}
    \mathcal{L}_g h_{proj} = \begin{bmatrix}
        < \prel + \vrel \frac{\sqrt{\|\prel\|^2 - r^2}}{\|\vrel\|}, 
            \textbf{R}\begin{bmatrix}
                0  \\
                \frac{-Ll}{I_{xx}}\\
                \frac{1}{m}
            \end{bmatrix}>\\
                < \prel + \vrel \frac{\sqrt{\|\vrel\|^2 - r^2}}{\|\vrel\|}, 
            \textbf{R}\begin{bmatrix}
                \frac{Ll}{I_{yy}} \\
                0\\
                \frac{1}{m}
            \end{bmatrix}>\\
         < \prel + \vrel \frac{\sqrt{\|\vrel\|^2 - r^2}}{\|\vrel\|}, 
            \textbf{R}\begin{bmatrix}
                0  \\
                \frac{Ll}{I_{xx}}\\
                \frac{1}{m}
            \end{bmatrix}>\\
         < \prel + \vrel \frac{\sqrt{\|\vrel\|^2 - r^2}}{\|\vrel\|}, 
            \textbf{R}\begin{bmatrix}
                \frac{-Ll}{I_{yy}} \\
                0 \\
                \frac{1}{m}
            \end{bmatrix}>
    \end{bmatrix}^T,
\end{align}
Using the same arguments we gave in the proof of theorem. \ref{thm:CC-CBF-3D}, we can infer that $\mathcal{L}_gh_{proj}$ cannot be zero. This implies that $h_{proj}$ is a valid CBF.
\end{proof}

\begin{remark}
{\it

Based on Theorems \eqref{thm:CC-CBF-3D} and \eqref{thm:CC-CBF-proj}, where $\mathcal{L}_g h \neq 0$, we can utilize the conclusion from \cite[Theorem 8]{XU201554} to deduce that the control inputs obtained from the resulting CBF-QP \eqref{eqn: CBF-QP} are Lipschitz continuous. As a result, 
the resulting solutions guarantee forward invariance of the safe set generated by the proposed C3BF candidates.
}
\end{remark}

\subsection{Point mass model}
\label{section: point model cccbf}
\par Similar to Theorem \ref{thm:bicycletheorem}, we can establish similar results for simple point mass models \cite{https://doi.org/10.48550/arxiv.2106.13176} of the form:

\begin{equation}
	\begin{bmatrix}
		\dot{p} \\
		\dot{v}
	\end{bmatrix}
	=
        \begin{bmatrix}
            0_{2x2} & I_{2x2} \\
            0_{2x2} & 0_{2x2}
        \end{bmatrix}
        \begin{bmatrix}
            p\\
            v\\
        \end{bmatrix}
	+
	\begin{bmatrix}
            0_{2x2} \\
            I_{2x2} 
	\end{bmatrix}
		u,
    \label{eqn:Acceleration controlled Point mass model}
\end{equation}
where $p$ = [$x_p$, $y_p$]$^T$, $v$ = [$v_x$, $v_y$]$^T$, and $u$ = [$a_x$, $a_y$]$^T$ $\in\mathbb{R}^2$ denotes the position, velocity and acceleration inputs, respectively. The proposed C3BF-QP is indeed a valid CBF, and its proof is straightforward.

\begin{figure}[t]
    \includegraphics[width=0.9\linewidth]{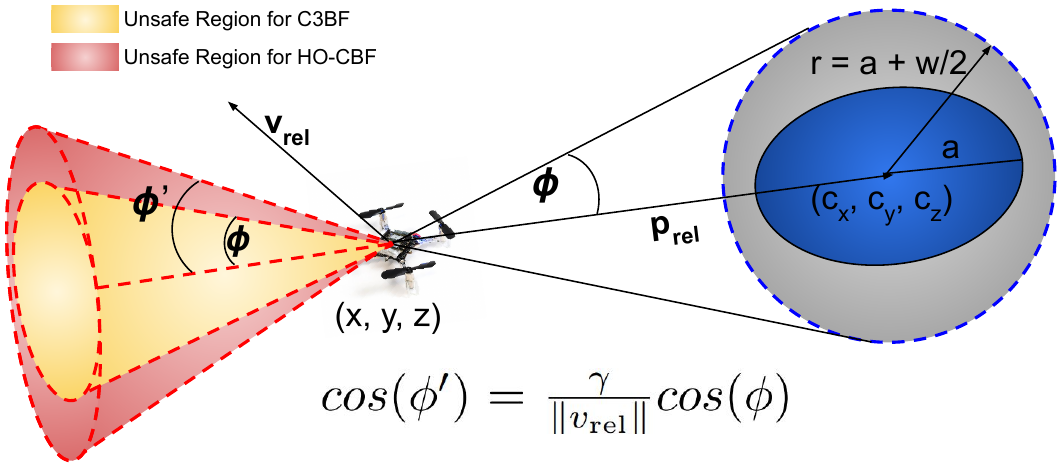}
\caption{Comparison of HO-CBF with C3BF. Here we are trying to compare the $\phi'$ and $\phi$ obtained from the two CBF formulations. It can be observed that $\phi'$ (pink cone) is dependent on $v_{rel}$, while $\phi$ (yellow cone) is a constant. The HO-CBF guarantees safety for a set that is not only smaller but also dependent on $v_{rel}$ as shown by the pink cone. Hence, HO-CBF is more conservative compared to C3BF.}
\label{fig:HO-C3-CBF}
\end{figure}

\subsection{Comparison with Higher Order CBFs}
We introduce the state-of-the-art Higher Order Control Barrier Functions (HO-CBFs) and compare with the proposed C3BF in this section. Since, the collision constraints are w.r.t. the position, the corresponding CBF is of relative degree two, we need to define a Higher Order CBF with $m = 2$ as in \cite[Eq 16]{9516971}, which is given as:
\begin{equation}
\begin{aligned}
\psi_{1}(x,t) &= \dot{b}(x,t) + p\alpha_{1} (b(x,t))\\
\psi_{2}(x,t) &= \dot{\psi_{1}}(x,t) + p \alpha_{2} (\psi_{1}(x,t)) ,\\
\end{aligned}
\end{equation}
where $b(\state,t) = (c_x(t) - x_p)^2 + (c_y(t) - y_p)^2 + (c_z(t) - z_p)^2 - r^2$, and r is the encompassing radius given by $r = max(c_1, c_2, c_3)$. $\alpha_1, \alpha_2$ are both class $\mathcal{K}$ functions, and $p$ is a tunable constant. As explained previously, $c_x,c_y,c_z$ is the centre location of the obstacle as a function of time. Let us examine the form of HO-CBF where $\alpha_{1}$ is a square root function (which is also strictly increasing), $\alpha_{2}$ is a linear function, due to its similarity to C3BF. Consequently, the resulting Higher Order CBF candidate takes the following form:

\begin{equation}
    h_{HO}(\state, t) = < \prel, \vrel> + \gamma \sqrt{(\|\prel\|^2 - r^2)}.
    \label{eqn:HO-CBF}
\end{equation}
%
We can show that the above mentioned HO-CBF is also a valid CBF as we did in Theorems \ref{thm:CC-CBF-3D} and \ref{thm:CC-CBF-proj}. We will now compare it with the proposed C3BF.

The C3BF concept aims to prevent the $\vrel$ vector, which represents the relative velocity between the quadrotor and the obstacle, from entering the collision cone region defined by the half-angle $\phi$. Figure \ref{fig:HO-C3-CBF} illustrate this idea. We can rewrite the HO-CBF formula presented in \eqref{eqn:HO-CBF} in the following form: 
\begin{align}
h_{HO}(\state, t) = < \prel, \vrel> + \|\prel\| \|\vrel\| cos(\phi')
    \label{eqn:HO-CBF-CC}
\end{align}
where, $cos(\phi') = \frac{\gamma}{\|\vrel\|}cos(\phi)$.
If we are able to identify a suitable $\gamma$ (penalty term) for the given HO-CBF, it would result in a valid CBF as per \cite{9516971}. Nonetheless, in such a scenario where $\gamma$ remains constant and $\|\vrel\|$ goes on increasing, it leads to increase in $\phi'$, thus, overestimating the cone as can be seen in Fig.\ref{fig:HO-C3-CBF}. Conversely, with the C3BF approach, we permit the penalty term to vary over time, i.e., $\gamma = \|\vrel\|$, resulting in a more precise estimation of the collision cone compared to the HO-CBF case. 
This also shows that C3BF is not a special case of Higher Order CBF.
This is also evident from the simulation outcomes of both CBFs, as demonstrated in Section \ref{section: Results}.

\section{Results and Discussions}
\label{section: Results}
\par In this section, we provide the simulation and hardware experimental results to validate the proposed C3BF-QP. All the simulations are done in Pybullet \cite{coumans2019}, a python-based physics simulation engine, on a computer with Ubuntu 22.04 with Intel i7-12800 HX CPU, with 16GB RAM and Nvidia 3070Ti GPU. 
For the class $\mathcal{K}$ function in the CBF inequality, we chose $\kappa(h) = \gamma h$, where $\gamma=1$.




\subsection{Unicycle Model}

We have considered the reference control inputs as a simple PD-controller.
We chose constant target velocities for verifying the C3BF-QP. However, the reference controller can be replaced by any trajectory tracking controller like the Stanley controller \cite{4282788}. The QP yields the optimal accelerations, which are then applied to the robot.

\textit{Simulation Results:}
All the simulations were performed on TurtleBot3 UGV in Pybullet. We consider different scenarios with different poses and velocities of both the vehicle and the obstacle. Different scenarios include static obstacles resulting in a) turning, b) braking and moving obstacles resulting in c) reversing, and d) overtaking.

\begin{figure}
    \centering
    \includegraphics[width=1\linewidth]{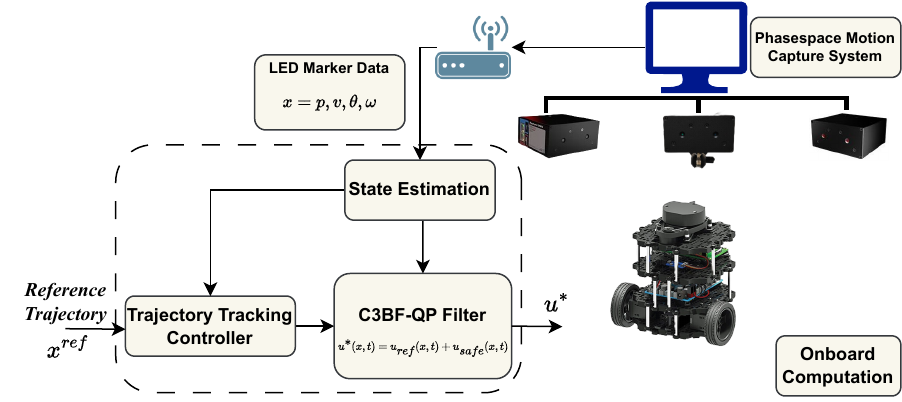}
\caption{Experimental Setup for TurtleBot Burger.} 
\label{fig: Experimental_setup_unicycle}
\end{figure}

\textit{Hardware Results:}
The experiments were performed on TurtleBot3 Burger (Fig. \ref{fig:turtlebot stoch-jeep and CF}), a two-wheeled differential drive type platform, to demonstrate the efficacy of C3BF on Unicycle models.
The global position of the car, as well as the obstacle(s), is measured using PhaseSpace\textsuperscript{\texttrademark} motion capture system with a tracking frequency of 960 $Hz$. The 2 LED Markers are placed in front and rear of the bot to estimate its state ($p, v, \theta, \omega$). The experimental setup is shown in Fig \ref{fig: Experimental_setup_unicycle}.

\begin{figure}
   \centering
    \includegraphics[width=0.4\textwidth]{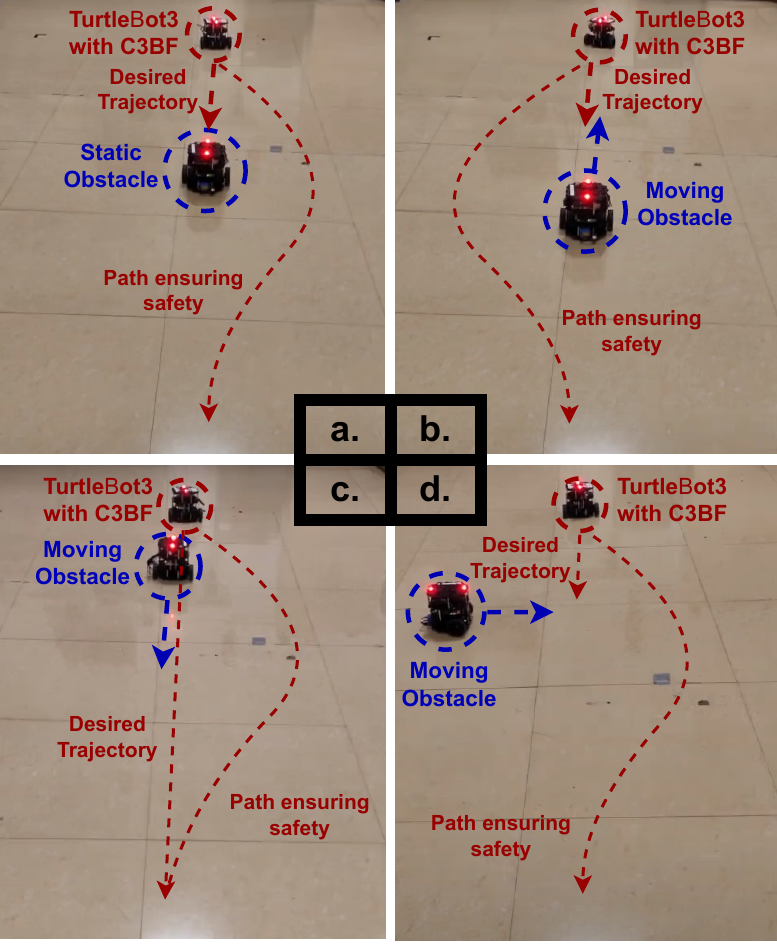}
    \caption{Interaction of TurtleBot3 (Unicycle Model) with (a) Static Obstacle; (b) Dynamic Obstacle (Velocity towards the agent); (c) Dynamic Obstacle (Overtaking); (d) Dynamic Obstacle (Velocity perpendicular to \& towards the agent).}
    \label{fig:unicycle_hardware}
\end{figure}

Similar to the experiments conducted in the simulations, we observed braking, turning, and overtaking behaviors. Irrespective of the initial conditions of the robot and the obstacle, collision is avoided in all cases as shown in Fig \ref{fig:unicycle_hardware}.

\subsection{Bicycle Model}
We have extended and validated our C3BF algorithm for the bicycle model \eqref{eqn:bicyclemodel with small beta} which is a good approximation of actual car dynamics under the assumption of small lateral acceleration ($\leq 0.5\mu g$, $\mu$ is the friction co-efficient) \cite{https://doi.org/10.48550/arxiv.2103.12382} \cite{7995816}.  
The linear acceleration reference control $a_{ref}$ was obtained from a PD-controller tracking the desired velocity, and the steering reference $\beta_{ref}$ was obtained from the Stanley controller.

\textit{Simulation Results:}
All the simulations were performed on the bicycle model in PyBullet. 

\textit{Hardware Results:}
Corresponding to the cases tested in Python simulation, experiments were performed on Stoch-Jeep (Fig. \ref{fig:turtlebot stoch-jeep and CF}), a rear wheel drive car $1/10th$ scaled model, which is powered by 11.1 \textit{V}, 1800 \textit{mAh} LIPO Battery. The steering actuation is handled by a servo motor TowerPro MG995, and a DC motor actuates the rear wheels.
The global position of the car, as well as the obstacle(s), is again measured using PhaseSpace\textsuperscript{\texttrademark} motion capture system with a tracking frequency of 960 $Hz$. The 2 LED Markers are placed in front and rear of the car to estimate the state of the car ($p, v, \theta, \omega$).

\begin{figure}
    \centering
    \includegraphics[width=1\linewidth]{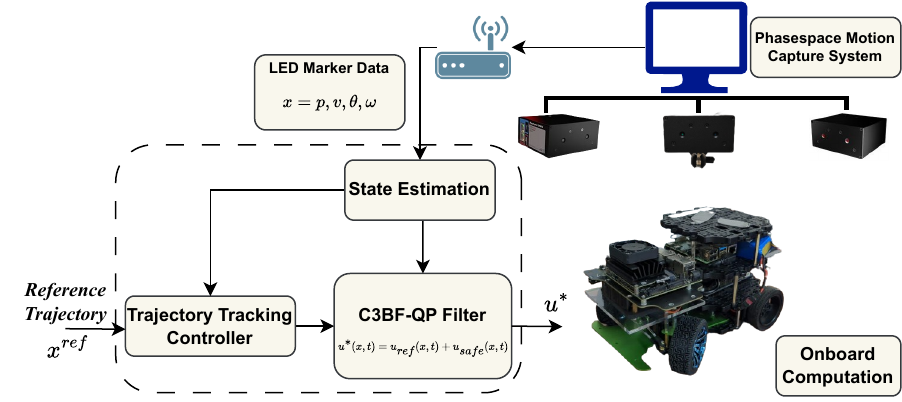}
\caption{Experimental Setup for Stoch-Jeep.} 
\label{fig: Experimental_setup_bicycle}
\end{figure}

\begin{figure}
   \centering
    \includegraphics[width=0.4\textwidth]{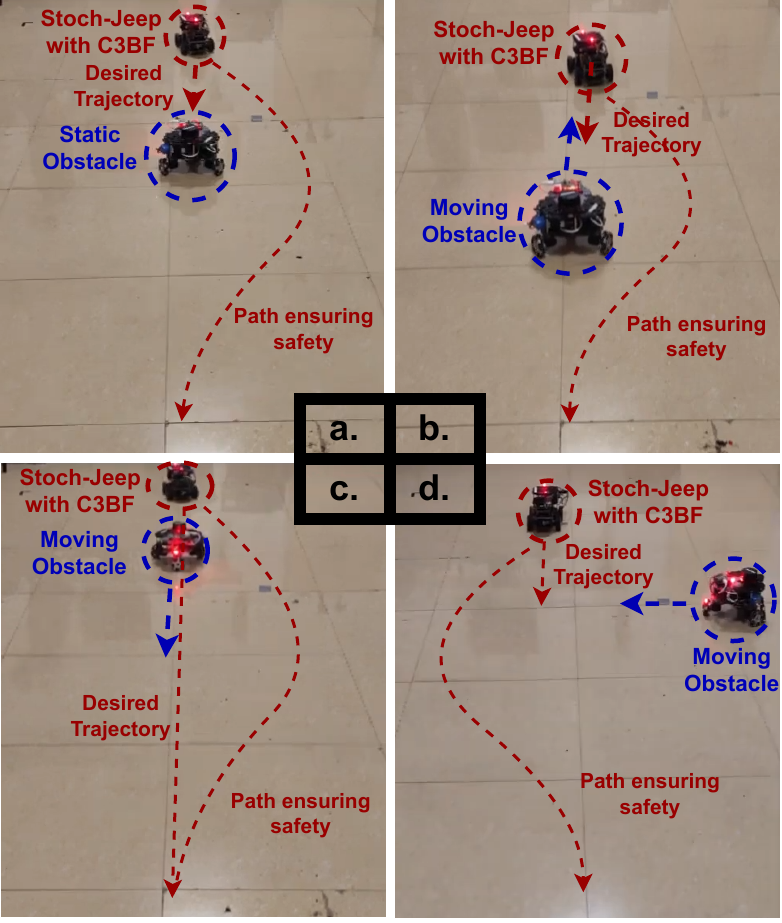}
    \caption{Interaction of Stoch-Jeep (Bicycle Model) with (a) Static Obstacle; (b) Dynamic Obstacle (Velocity towards the agent); (c) Dynamic Obstacle (Overtaking); (d) Dynamic Obstacle (Velocity perpendicular to \& towards the agent).} 
    \label{fig:bicycle_hardware}
\end{figure}
Similar to the experiments conducted in the simulations, we observed braking, turning, and overtaking behaviors. Irrespective of the initial conditions of the robot and the obstacle, collision is avoided in all cases.

\subsection{Quadrotor Model}
We have validated the C3BF-QP based controller on quadrotors for both 3D and Projection CBF cases. Again, PD Controller is used as a reference controller to track the desired path, and the safety controller deployed is given by Section \ref{subsection: safe_controller}. We chose constant target velocities for verifying the C3BF-QP. 

\begin{figure}
    \centering
    \includegraphics[width=1\linewidth]{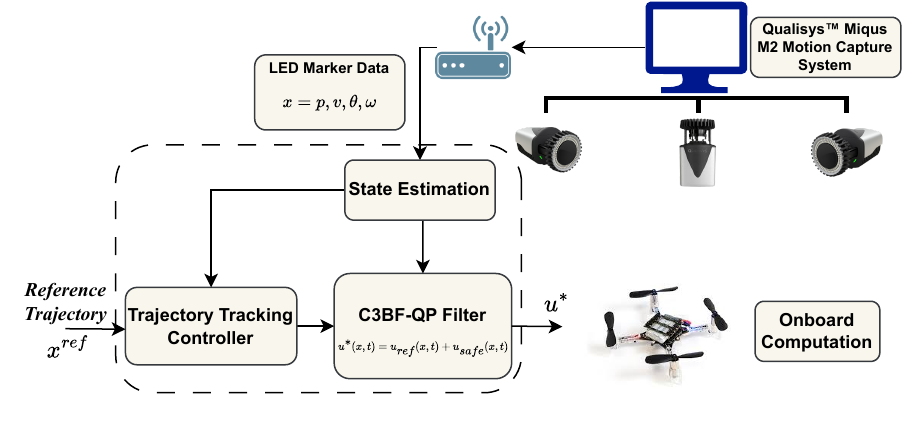}
\caption{Experimental Setup for Crazyflie 2.1.} 
\label{fig: Experimental_setup_UAV}
\end{figure}

\begin{figure}
   \centering
    \includegraphics[width=0.4\textwidth]{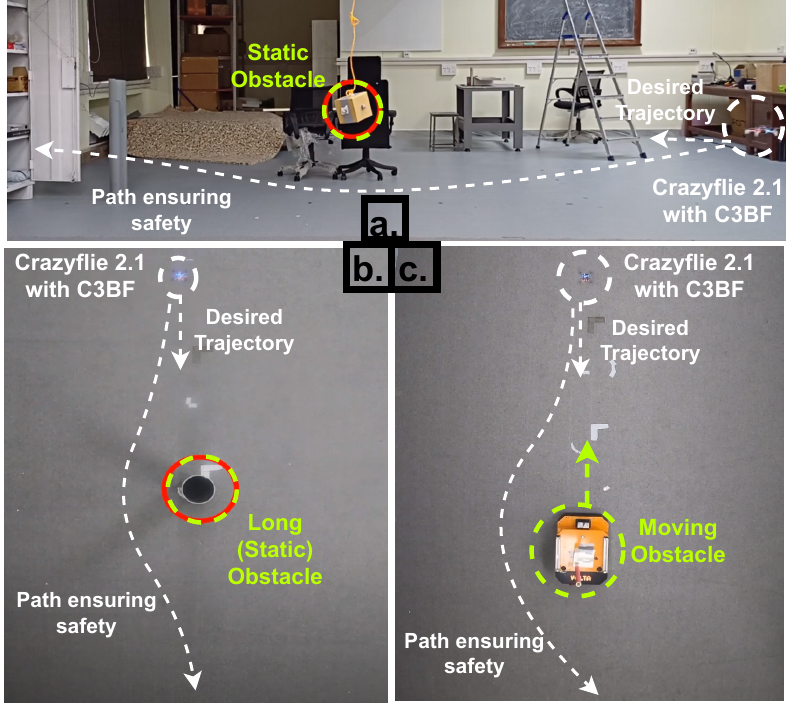}
    \caption{Interaction of Crazyflie 2.1 (Quadrotor) with (a) Static Obstacle; (b) Long (Static) Obstacle; (c) Dynamic Obstacle (Velocity towards the agent).}
    \label{fig:uav_hardware}
\end{figure}

\textit{Simulation Results:}
 The simulations were conducted using the multi-drone environment \cite{pybullet-drones} on Pybullet. 

\textit{Hardware Results:}
\begin{figure*}
       \centering
        \begin{subfigure}[b]{0.32\textwidth}
        \includegraphics[width=\textwidth]{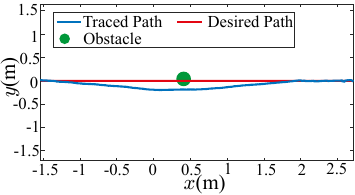}
        \caption{}
        \label{fig:single_cyl_path}
        \end{subfigure}
        \begin{subfigure}[b]{0.32\textwidth}
        \includegraphics[width=\textwidth]{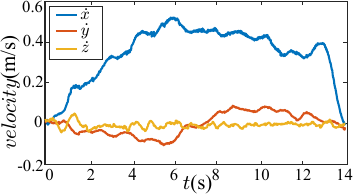}
        \caption{}
        \label{fig:single_cyl_vel}
        \end{subfigure}
        \begin{subfigure}[b]{0.32\textwidth}
        \includegraphics[width=\textwidth]{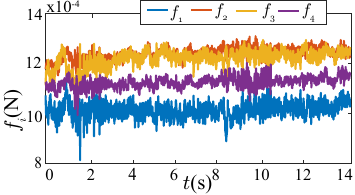}
        \caption{}
        \label{fig:single_cyl_inp}
        \end{subfigure}
        \caption{Experimental results with a single cylindrical obstacle. a) Traced Path. b) Evolution of linear velocity. c) Evolution of control inputs.}
        \label{fig:single_cyl_exp}
    \end{figure*}

    \begin{figure*}
       \centering
        \begin{subfigure}[b]{0.32\textwidth}
        \includegraphics[width=\textwidth]{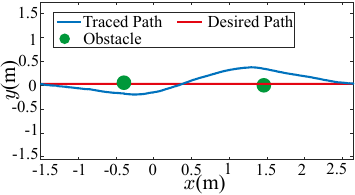}
        \caption{}
        \label{fig:double_cyl_path}
        \end{subfigure}
        \begin{subfigure}[b]{0.32\textwidth}
        \includegraphics[width=\textwidth]{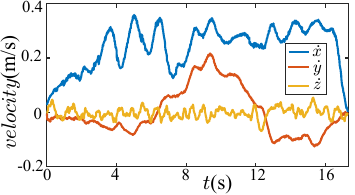}
        \caption{}
        \label{fig:double_cyl_vel}

        \end{subfigure}
        \begin{subfigure}[b]{0.32\textwidth}
        \includegraphics[width=\textwidth]{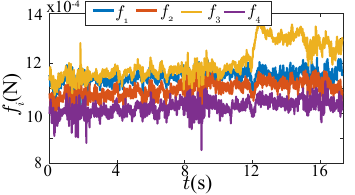}
        \caption{}
        \label{fig:double_cyl_input}
        \end{subfigure}
        \caption{Experimental results with a two cylindrical obstacle. a) Traced Path. b) Evolution of linear velocity. c) Evolution of control inputs.}
        \label{fig:double_cyl_exp}
    \end{figure*}

    \begin{figure*}
       \centering
        \begin{subfigure}[b]{0.32\textwidth}
        \includegraphics[width=\textwidth]{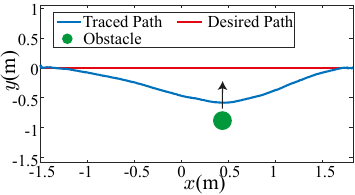}
        \caption{}
        \label{fig:moving_obs_path}
        \end{subfigure}
        \begin{subfigure}[b]{0.32\textwidth}
        \includegraphics[width=\textwidth]{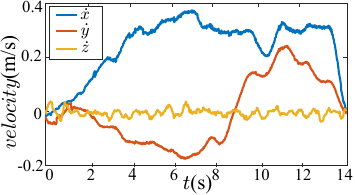}
        \caption{}
        \label{fig:moving_obs_vel}
        \end{subfigure}
        \begin{subfigure}[b]{0.32\textwidth}
        \includegraphics[width=\textwidth]{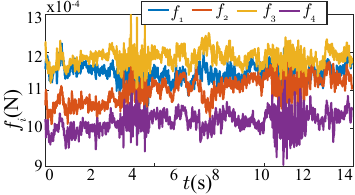}
        \caption{}
        \label{fig:moving_obs_inp}
        \end{subfigure}
        \caption{Experimental results with a moving obstacle. a) Traced Path. b) Evolution of linear velocity. c) Evolution of control inputs.}
        \label{fig:moving_obs_expt}
    \end{figure*}

The experiments were conducted using the Bitcraze\textsuperscript{\texttrademark} Crazyflie 2.1 aerial drone to demonstrate the efficacy of the C3BF-QP controller framework. 
The global position of the drone as well as the obstacle is measured using Qualisys\textsuperscript{\texttrademark} Miqus M3 motion capture system with a tracking frequency of 100 $Hz$. Further, for the drone, the global position from the motion capture system are fused with the onboard IMU data via Extended Kalmann Filter to the get the filtered state.The control commands are generated by an off board computer and transmitted to the drone via a radio link. The communication with the drone is facilitated through the Crazyflie Python library \cite{cfclient}. Experiments are performed for the cases with single static obstacle, multiple static obstacles and moving obstacles. The corresponds results are shown in Figs.~{\ref{fig:single_cyl_exp}-\ref{fig:moving_obs_expt}}. As is observed from Figs.~{\ref{fig:single_cyl_path}-\ref{fig:moving_obs_path}}, the quadrotor is able to successfully evade the obstacles. The corresponding command inputs and the resultant linear velocities are shown in Figs.~{\ref{fig:single_cyl_inp}-\ref{fig:moving_obs_inp}} and Figs.~{\ref{fig:single_cyl_vel}-\ref{fig:moving_obs_vel}}, respectfully. Hence, the experimental results verify the efficacy of the proposed scheme for obstacle avoidance.

\subsection{Comparison between C3BF and HO-CBF}
All the aforementioned cases were tested with the HO-CBF to compare its performance against C3BF. We observe that the HO-CBF could not avoid a high-speed approaching obstacle. Moreover, it is not able to properly avoid the longer obstacles in the projection CBF case. These shortcomings of the Higher Order CBF are demonstrated in the supplementary video.

\subsection{Robustness of C3BF}
Without changing the above control framework we can observe that the C3BF is robust in the following two cases: 

\subsubsection{Multiple Obstacles}
\begin{figure}
   \centering
    \includegraphics[width=0.4\textwidth]{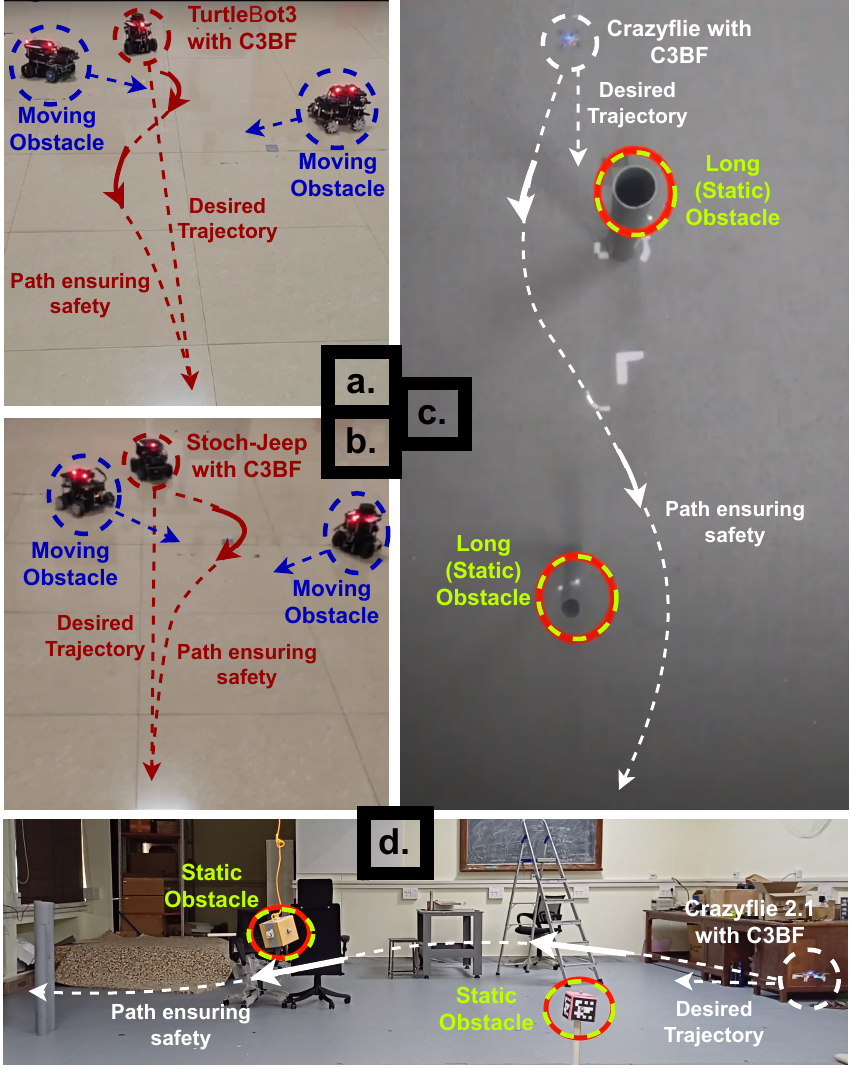}
    \caption{Robustness in scenarios with multiple obstacles: (a) TurtleBot3 (Unicycle Model) [Dynamic Obstacles]; (b) Stoch-Jeep (Bicycle Model) [Dynamic Obstacles]; (c) Quadrotor (Crazyflie 2.1) - Long (Static) Obstacles; (d) Quadrotor (Crazyflie 2.1) - Spherical (Static) Obstacles}
    \label{fig:robustness_multiobst}
\end{figure}

The vehicles successfully navigates through a series of obstacles (both Spherical and Long obstacles) avoiding collisions and showcasing robustness as can be observed in Fig. \ref{fig:robustness_multiobst}

\subsubsection{Multiple agents with C3BF-QPs}
In multi-agent scenarios, where multiple vehicles employ the collision cone CBF-QP, both the ego-vehicle and the approaching vehicle are able to avoid collision in different configurations (static or moving), thus demonstrating robustness with respect to obstacles following the same Collision Cone CBF controller.

All the simulations and hardware experiments can be viewed in the attached video link\footnote{\label{note: Webpage}.\url{https://tayalmanan28.github.io/CollisionConeCBF/}}


\section{Conclusions}
\label{section: Conclusions}
We presented a novel real-time control methodology for Unmanned Ground Vehicles (UGVs) and Unmanned Aerial Vehicles (UAVs) for avoiding moving obstacles by using the concept of collision cones. The combination of collision cones and CBFs gives an ability to handle moving obstacles and guarantees collision avoidance. 
We successfully constructed CBF-QPs with the proposed CBF for the unicycle, bicycle and quadrotor models and demonstrated the framework both on simulations and robot hardware.
We also showed that the 
current state-of-the-art Higher Order CBFs is more conservative and fails in certain scenarios. 
Finally, we demonstrated the robustness of the proposed CBF-QP controller with respect to safe navigation in a cluttered environment consisting of multiple obstacles and agents with the same safety filters.

In our future work, we plan to combine this control framework with Model Predictive Control (MPC), Reinforcement Learning (RL) based approaches, on more complex models like quadrupeds.  

\label{section: References}
\bibliographystyle{IEEEtran}
\bibliography{references.bib}

\end{document}